%% file: main_abstract.tex
\documentclass[11pt]{wlscirep}
\usepackage[utf8]{inputenc}
\usepackage[T1]{fontenc}
\usepackage{hyperref}
\usepackage{bm}
\usepackage{epstopdf}
\usepackage{makecell}
\usepackage{graphicx}
\usepackage{lscape}
\usepackage{pdflscape}
\usepackage{multirow}
\usepackage{amsmath}
\usepackage{makecell}
\usepackage{filecontents}
\usepackage{epsfig}
\usepackage{amssymb}
\usepackage{threeparttable}  
\usepackage{float}
\usepackage{epsfig}
\usepackage{balance}
\usepackage{booktabs}
\usepackage{enumitem}
\usepackage{mathtools}
\usepackage{makecell}
\usepackage{multirow}
\usepackage{pifont}
\usepackage{stackengine}
\usepackage{xcolor}
\usepackage{colortbl}
\usepackage{tabu}
\usepackage{geometry}
\usepackage{fancyhdr}
\usepackage{array}
\usepackage{subcaption}
\usepackage{graphicx}
\usepackage{caption}
\usepackage{color}
\usepackage{xspace}
\usepackage[breakable]{tcolorbox}
\usepackage{titletoc}
\usepackage{setspace} % Include the setspace package
\usepackage{diagbox}
\usepackage{caption}
\usepackage{subcaption}
\usepackage{lineno}
\usepackage{ccaption}
\usepackage{capt-of}
\usepackage{graphicx}

\usepackage[backend=biber,style=nature,sorting=none,url=false,maxnames=99,minnames=1]{biblatex}
\DeclareBibliographyCategory{sec1}
\addtocategory{sec1}{report, islam2024calibrated, bougna2022quantitative, wen2021applications, mannering2020big, yan2023learning, tazul2017understanding, mannery2014analytic, dong2018improved, theofilatos2019comparing, rahim2021deep, sattar2023transparent, karimi2023crashformer, SHARMA2020100127, xu2019statistical, MA2021106322, boggs2020exploratory, wali2020relationship, ahmed2021correlated, liu2023tap, lu2022learning, hu2021lora, gao2023s, fhwa_vision_zero, rezapour2022identification, hsis, googlemaps, police_crash_reports, crashworthiness_data, abdel2005analysis, IRANITALAB201727, SAVOLAINEN20111666, bordt2023shapley, shapley:book1952, ghorbani2019data, jia2019towards, meta2024llama, zhao2024surveylargelanguagemodels, sanh2022multitaskpromptedtrainingenables, longpre2023flancollectiondesigningdata, dai2023saferlhfsafereinforcement, bai2022traininghelpfulharmlessassistant, shapley1953value, zhang2024vision, bhuiyan2022crash,carrodano2024data,illinois_bac_limit,washington_bac_limit, 10.1145/3588728,randomforest, FREUND1997119, prokhorenkova2019catboost, quinlan1986induction, cox1958regression, 10.1145/2939672.2939785, nationalbaseline,abdel2024matched,mannering2014analytic,achiam2023gpt,niu2024cross, zhen2024leveraging, jaradat2024multitask, de2023llm}

\DeclareBibliographyCategory{sec2}
\addtocategory{sec2}{du2024advancing, ahmed2023study, freund1999short, prokhorenkova2018catboost, chen2022explaining, chen2023algorithms, NIPS2017_8a20a862, kokalj-etal-2021-bert,goldshmidt2024tokenshap}
\title{\fontsize{16pt}{14pt}\selectfont Towards Reliable and Interpretable Traffic Crash Pattern Prediction and Safety Interventions Using Customized Large Language Models}

\newcommand{\be}{\begin{equation}}
\newcommand{\ee}{\end{equation}}
\newcommand{\ba}{\begin{align}}
\newcommand{\ea}{\end{align}}
\newcounter{extfig}

\def\ba{{$\bm{a}$}}

\def\trafficsafe{\textit{TrafficSafe}\xspace}
\def\dataset{\textit{TrafficSafe Event} dataset\xspace}
\def\model{\textit{TrafficSafe LLM}\xspace}
\def\attribution{\textit{TrafficSafe Attribution}\xspace}

\definecolor{myblue}{RGB}{0, 133, 186}
\definecolor{mygreen}{RGB}{72, 194, 0}
\definecolor{myred}{RGB}{177, 0, 28}
\definecolor{good}{rgb}{0.11, 0.77, 0.11}
\definecolor{bad}{rgb}{0.77, 0.11, 0.11}

\author[1, 2+]{Yang Zhao}
\author[1, 2+]{Pu Wang}
\author[1, 2]{Yibo Zhao}
\author[1, 2]{Hongru Du}
\author[1, 2*]{Hao (Frank) Yang}

\affil[1]{Center for Systems Science and Engineering, Johns Hopkins University,
Baltimore, MD, USA.}
\affil[2]{Department of Civil and Systems Engineering, Johns Hopkins University, Baltimore, MD, USA.}

\small{\affil[+]{The authors contributed equally.}
\affil[*]{The corresponding authors information:~haofrankyang@jhu.edu}}

{ % Set line spacing to 1.5
\nolinenumbers \begin{abstract}
\fontsize{11pt}{12pt}\selectfont

Predicting crash events is crucial for understanding crash distributions and their contributing factors, thereby enabling the design of proactive traffic safety policy interventions. However, existing methods struggle to interpret the complex interplay among various sources of traffic crash data, including numeric characteristics, textual reports, crash imagery, environmental conditions, and driver behavior records. As a result, they often fail to capture the rich semantic information and intricate interrelationships embedded in these diverse data sources, limiting their ability to identify critical crash risk factors. In this research, we propose \trafficsafe, a framework that adapts Large Language Models (LLMs) to reframe crash prediction and feature attribution as text-based reasoning. A multi-modal crash dataset including 58,903 real-world reports together with belonged infrastructure, environmental, driver, and vehicle information is collected and textualized into \dataset (totaling 12.74 million words). By customizing and fine-tuning state-of-the-art LLMs on this dataset, the proposed \model achieves a 42\% average improvement in F1-score over baselines across multiple crash prediction tasks, particularly for severe crashes. To interpret these predictions and uncover contributing factors, we introduce \attribution, a sentence-level feature attribution framework enabling conditional risk analysis. Findings show that alcohol-impaired driving is the leading factor in severe crashes, with aggressive and impairment-related behaviors having nearly twice the contribution for severe crashes compared to other driver behaviors. In addition, the co-occurrence of crash-contributing factors, such as alcohol-impaired driving, work zones, improper driving behaviors and other factors can significantly elevate risk levels. Furthermore, \attribution highlights pivotal features during model training, guiding strategic crash data collection for iterative performance improvements. The proposed \trafficsafe offers a transformative leap in traffic safety research based on foundation models, providing a blueprint for translating advanced artificial intelligence technologies into responsible, actionable, and life-saving outcomes. It is now reshaping how traffic researchers and policymakers approach the road safety. %To the best of the authors’ knowledge, this is the first multi-modal traffic crash analysis and feature attribution framework based on LLMs.

\end{abstract}

\begin{document}

% \sectiontexcount{introduction}
% \sectiontexcount{Novelties&Contributions}
% \sectiontexcount{results}
% \sectiontexcount{discussion}

% Task List:
% \begin{itemize}
%     \item add citations/ nature references Yang, Pu
%     \item code, Yang, Pu
%     \item Grammar tenses need to be consistent, present tense or past tense @all
% \end{itemize}

% \newpage

\flushbottom
\maketitle
% \thispagestyle{empty}
% \linenumbers
\input{introduction}

\input{Novelties_Contributions}

\input{results}

\input{discussion}
\input{conclusion}
\printbibliography[title={Reference}, category=sec1]
\newpage 
% \bibliography{Main}
% \bibliographystyle{nature}

\input{method}
\printbibliography[title={Reference}, category=sec2]
\newpage 

\section{Data Availability}
Details of each raw data source and data processing are described in the Method Section. The processed data examples are available at \url{https://github.com/Puw242/TrafficSafe}. In compliance with HSIS data policy, requests for the complete raw dataset should be made via \url{https://highways.dot.gov/research/safety/hsis}. 
\section{Code Availability}
Code is publicly accessible at \url{https://github.com/Puw242/TrafficSafe}.
% \input{data_availability}
% \input{code_availability}
% \bibliographystyle{nature}
% \bibliography{method}

% \input{Acknowledgement}
% \input{Author_Contributions}
% \input{Competing_interests}

%\section{Acknowledgment}
%This research is supported by XXX.

\section{Author Contributions}
Y.Z., P.W. and H.F.Y conceptualized and designed the study. P.W. and Yibo Z. collected data. P.W. and Yibo Z. processed the data and designed prompts. Y.Z. and P.W. performed experiments. Yibo Z. run the baseline models. Y.Z., P.W., Yibo Z. and H.F.Y prepared the figures. Y.Z., P.W., Yibo Z. and H.F.Y analyzed results. Y.Z., P.W., Yibo Z. and H.F.Y wrote the initial draft. H.D. and H.F.Y provided guidance and feedback for the study. H.D. and H.F.Y revised the manuscript. H.F.Y. acquired the funding. H.F.Y. provided computational resources. All authors prepared the final version of the manuscript. 

\section{Competing Interests}
The authors declare no competing interests.

\newpage \input{extended_data}

\end{document}

%% file: introduction.tex
% I. Introduction: 800 words
% 1. problem less than 8 sentences one paragraph
% The first sentence is summary. (1) this problem is important (2) this is difficult, a lots of factors, multi-modal data, current method not in event level

% 2. Exsiting literature summary. Not critisize. 6 W. Chanllenges

% 3. What's our thoughts. Current tech how to support our idea. LLM strong ability, causality Computation improvements. Detail

% 4. Detail of our method

% II Contribution Novelty  200 words

% summary

% 1. dataset
% 2. model. Robust result. 
% 3. causality. Training: better data selection. Inference: features importance
% 4. Overall framework. Safety policy intervension, findings, etc.

% III Result 2000 words
% data and model description

% data preparation input and output 300

% 700 words. event-level prediction result. 1. results, 2. confidence/defelity, 3. work in both states
% 1500 words. interpretability 1. data contribution 2. causal analysis. New Figrue

% 1. spatial regions, temperal analysis, large-scale 

% IV Shapley analysis

% urban rural, speed limit, crash type, sev fatal, pedestrain cyclist, waether, infrastructure

% spatial-temperal information, Infrastureture Information, Dynamic Conditions, Unit& person

% one paragrapgh for IL and WA

% V. Discussion  500 words

% Sev: serious and fatal for each State. Merge positive and negative together
% Type: 3 Figure 
\vspace{-2em}
\section{Introduction}

\begin{figure}[!p]
    \centering
    \includegraphics[width=.93\linewidth]{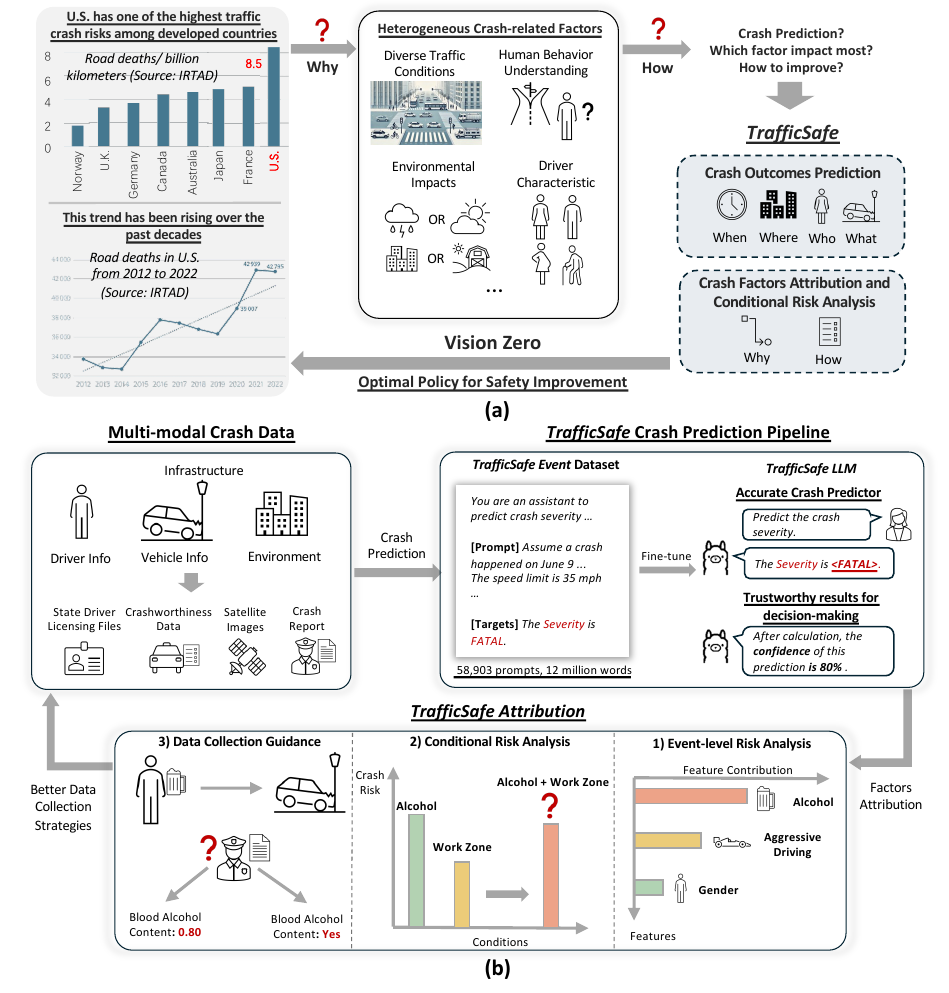}
    \vspace{-10pt}
    \caption{\small \textbf{Overview of the Proposed \trafficsafe Framework}. \textbf{(a)} The U.S. faces one of the highest crash risks among developed countries, with a rising trend. However, analyzing and addressing this issue is challenging due to the heterogeneous factors involved in crash events, including traffic conditions, human behavior, environmental impacts, and driver characteristics. To tackle this, we propose \trafficsafe, a framework designed for two key tasks: \textbf{1) Predicting crash outcomes} and \textbf{2) Attributing crash factors with conditional risk analysis}. By addressing questions such as why crashes occur and how to mitigate crash risks, \trafficsafe seeks to deliver optimal policy for safety improvement, aligning with the Vision Zero goal \supercite{fhwa_vision_zero}. \textbf{(b)} The \trafficsafe workflow incorporates multi-modal data, including driver behavior, vehicle details, infrastructure, and environmental conditions, represented through textual reports, satellite imagery, and other formats. Leveraging an AI-expert cooperative method, the crash data is transformed into textual prompts, resulting in the \dataset comprising 58,903 prompts. \model is created with accurate and trustworthy forecasting abilities for further analysis. Building on this pipeline, \attribution operates across three dimensions: 1) Event-level risk analysis to identify feature contributions, 2) Conditional risk analysis to assess state-level risks under varying conditions, and 3) Data collection guidance to optimize the data acquisition process. The results of \attribution provide actionable insights to enhance data analysis and collection, fostering a more comprehensive understanding of crash data and events.}
    \label{fig:1}
\end{figure}

Predicting traffic crash outcomes at the event level can greatly improve our understanding of crashes contributing factors and support the safety policy interventions. Currently, the United States has one of the highest traffic crash risks among developed countries (see Figure \ref{fig:1}a), with 42,795 fatalities reported in 2022 \supercite{report}. The number of fatalities still shows a persistent upward trend over recent decades, highlighting the urgent need for innovative approaches to uncover the major causes of crashes and provide actionable insights for policy interventions. An effective data-driven crash prediction model can learn from historical crash-related data, and offer potential guidance for reducing crash risk by identifying the leading factors of crashes \supercite{islam2024calibrated}. Current research for crash prediction can be grouped into two groups: 1) macroscopic (statistic-level) prediction \supercite{bougna2022quantitative, wen2021applications, mannering2020big, yan2023learning, tazul2017understanding} and 2) microscopic (event-level) prediction \supercite{mannering2014analytic, dong2018improved, theofilatos2019comparing, rahim2021deep, sattar2023transparent}. Macroscopic prediction typically relies on statistical methods to gain a general understanding of safety levels, compare the safety performance of different areas and time frames, identify high-risk zones, and track safety trends \supercite{bougna2022quantitative, wen2021applications}. While these methods can partially predict \textit{when} and \textit{where} crashes are more likely to occur, they fail to forecast \textit{who} is involved, \textit{what} types will likely to be, \textit{why} crashes happen, and \textit{how} to mitigate risks at event granularity \supercite{mannering2020big, carrodano2024data}. To address this limitation, microscopic crash prediction, which focuses on specific traffic conditions and circumstances, has been developed to predict the crashes consequences using machine learning (ML) approaches \supercite{theofilatos2019comparing, sattar2023transparent}. Despite their potential in answering \textit{who} and \textit{what}, these models face limitations in crash prediction precision and generalization \supercite{wen2021applications, mannering2020big}. Moreover, integrating multi-modal traffic crash data and interpreting model's outputs (together with contributing factors) remain challenging. Consequently, existing crash prediction models struggle to accurately forecast crash outcomes and effectively incorporate their insights into the design of policy intervention.

Crash data are inherently heterogeneous, making accurate prediction a significant challenge. After the crash happened, first responders compile textual and numerical on-site details, often supplemented by images, driver behavior data, and licensing records. Although these diverse sources hold immense potential for crash prediction and feature attribution, three key obstacles must be addressed: \textbf{1) Data Integration.} Existing approaches often reduce multi-modal data to one-hot embeddings for classification tasks \supercite{SHARMA2020100127, xu2019statistical, abdel2024matched}. However, these approaches often neglect the valuable information contained within textual and behavioral data, potentially limiting the accuracy and reliability of crash prediction models \supercite{theofilatos2019comparing, sattar2023transparent}. \textbf{2) Method Generalization.} Scaling crash-event prediction models to new data remains a complex endeavor due to the large variety of features, the complexity of representation extraction and encoding, and the diverse formats in which crash data appear \supercite{boggs2020exploratory, wali2020relationship}. Current machine learning solutions are often tailored to specific data types, limiting their adaptability when new cases or additional data modalities arise \supercite{ahmed2021correlated, mannering2020big, liu2023tap, lu2022learning}. \textbf{3) Feature Learning and Attribution.} Multi-modal crash data regularly include partially overlapping information, such as road attributes recorded in both on-site images and textual crash reports, complicating the accurate assessment of each feature’s unique contribution.

Recent advancements in Large Language Models (LLMs), such as GPT-4 \supercite{achiam2023gpt} and LLaMA 3 \supercite{meta2024llama}, have demonstrated their potential for deriving complex crash patterns from multi-modal data \supercite{gao2023s} and addressing persistent challenges. However, fully adapting LLMs to predict crash outcomes and inform effective safety interventions requires overcoming three primary technical hurdles in data, model, and interpretability. From a data perspective, diverse crash records, including images, textual notes, and driver behavior logs, must be reformatted into textual inputs suitable for LLM processing. In terms of modeling, the generative nature of LLMs, which have extensive output vocabularies (e.g., LLaMA 3’s 128,256 tokens), poses challenges for discriminative learning tasks and raises concerns about trustworthiness, particularly when crash outcomes (e.g., crash type or severity) are well-defined by public agencies into finite categories. Furthermore, interpreting LLM's outputs for crash prediction becomes difficult, as it remains unclear how much we can trust the forecasting results and how individual factors contribute to crash outcomes. This lack of interpretability and robustness analysis hinders the development of data-driven, actionable plans for mitigating the crash risks. Recent emerging efforts, such as CrashLLM and other LLM-empowered traffic safety studies, have begun to address the mentioned first and second challenges by exploring the novel LLMs for crash risk prediction and analysis \supercite{fan2024learning, de2023llm, zhen2024leveraging}. However, these methods represent only initial attempts based on textual prompt engineering with limited safety knowledge adaptation. Crucially, they do not work on the interpretability gaps of LLM outputs, which are essential for decision support in answering critical \textit{why} and \textit{how} questions.

 This study advances traditional traffic safety analysis by shifting from aggregate-level considerations to event-level crash prediction. We propose \textbf{\trafficsafe} (see Figure \ref{fig:1}b), a novel LLM-driven framework designed for addressing these challenges to provide a comprehensive understanding of crash events. \trafficsafe comprises three main components: \dataset for multi-modal crash data integration, \model for crash outcomes prediction, and \attribution for conditional risk analysis. Together, these components enable accurate and trustworthy crash consequences prediction and risk attribution, answering the \textit{when}, \textit{where}, \textit{who}, \textit{what}, \textit{why}, and \textit{how} to support targeted traffic safety interventions. By reframing crash outcome prediction as a text-based reasoning task, \trafficsafe exploits the inherent language reasoning capabilities of LLMs to offer actionable insights for crash prevention, ultimately paving the way for data-driven safety solutions.

%% file: Novelties_Contributions.tex
\section{Novelties and Contributions}

This study advances traditional traffic safety analysis by shifting from aggregate-level considerations to event-level crash prediction. In particular, we customize LLMs to forecast expected crash consequences and attribute relevant features with enhanced accuracy and interpretability. Our proposed framework, \trafficsafe, supports reliable and accountable learning from multi-modal crash data, facilitating a deeper understanding of crash events. Key contributions and findings include:

%This research push the existing traffic safety analysis  into event level reformulating the crash prediction as a text reasoning problem, \trafficsafe leverages LLMs' extensive crash-related expertise and reasoning ability for crash events learning and understanding. This study contributes to the literature by:

\begin{itemize}
    \item \textbf{Unlocking multi-modal data integration and text reasoning for crash consequence prediction.} We introduce the \trafficsafe framework to extend the LLMs for crash outcomes prediction. Rather than treating crash features as isolated numerical inputs, \trafficsafe integrates them into the broader semantic context of traffic data. To effectively utilize and integrate the multi-modal crash data, the \dataset is constructed with 58,903 textual prompts totaling over 12 million words. The \model is then fine-tuned by framing crash outcome prediction as a task specific token generation task. This approach yields a 41.7\% increase in average F1-score across multiple crash consequence prediction tasks.
    
    % By doing so, \trafficsafe enables more meaningful interpretations of crash risk and unlocks new opportunities for data-driven insights and policy interventions.

    \item \textbf{Integrating traffic safety priors in LLMs for trustworthy crash predictions.} Compared with existing LLMs, we incorporate crash-domain knowledge and priors into the model’s vocabulary as special tokens. This addition allows us to tailor the output to specific crash categories, including crash type, severity, and number of injuries, thereby providing a direct way to measure the model’s trustworthiness and link to targeted interventions. Experimental results of \model show a strong correlation between increasing confidence in the model’s output and higher prediction accuracy, achieving over 70\% accuracy when the confidence score exceeds 60\%. Notably, the model reaches more than 95\% precision for fatal crash predictions when the confidence score surpasses 60\%. This feature offers quantitative evidence to support safety-oriented decision-making and helps close the gap regarding how to trust the model’s predictive results.

    %Compare with existing LLMs, we added crash domain knowledge and priors into LLM's vocabulary set as special tokens and then customized the outputs into expected crash categories, including crash type, severity and number of injuries, providing direct way to evaluate the trustworthiness. Experimental results show a strong correlation between higher confidence in \model's output and more reliable prediction accuracy, achieving over 70\% accuracy when the confidence score exceeds 60\%. Notably, the model delivers over 95\% precision for fatal crash predictions when the confidence score exceeds 70\%, even learned from 1\% of the total data samples. This feature provides the quantitative evidence to support safety-oriented decision-making, mitigting the gap that how can we trust the prediction result.

    %To effectively leverage the multi-modal crash data, We created the \dataset with 58,903 textual prompts and over 12 million words, then fine-tuned \model by framing crash outcome prediction as a specialized token generation task. This approach delivers a 41.7\% boost in average F1 scores across multiple tasks and shows that higher model confidence correlates with higher accuracy—underscoring its trustworthiness and potential for real-world decision making.
    
    \item \textbf{Advancing feature interpretation for conditional risk analysis and policy intervention, even in unseen scenarios.} The \attribution framework is proposed for conditional risk analysis, which is supported by a novel sentence-level feature contributions calculation method, enabling event-level feature attribution for textual inputs of \model. Then, the "what-if" conditional analysis can further identify and analyze the most critical risk conditions and their combinations. For instance, alcohol-impaired driving consistently emerges as a leading contributor to serious and fatal crashes. While driving in a work zone under sober conditions poses minimal risk, combining these conditions with alcohol consumption drastically increases danger, making it one of the most hazardous scenarios for severe crashes. Furthermore, aggressive and impairment-related behaviors demonstrate nearly double the impact on severe crashes compared to other driver behaviors. These insights lay the groundwork for implementing targeted traffic safety policies and interventions \supercite{rezapour2022identification}.

    \item \textbf{Guiding optimal data collection for efficient model evolution and lifelong learning.} A longstanding challenge in crash modeling is determining how to select valuable data from heterogeneous sources and prioritize which information first responders should capture during incident documentation. The proposed \attribution addresses this by estimating the contributions of multi-modal data during training, then quantifying which data types have the greatest impact on model performance. Such insights guide more effective traffic safety data collection, improving crash prediction accuracy while also supporting efficient, continuous model evolution through a targeted, data-driven strategy.

    %it becomes possible to design a more focused and strategic data collection procedure—one that emphasizes the most impactful sources. This, in turn, enhances crash prediction accuracy and supports efficient, continuous model evolution. The \attribution framework can identify which data types contribute most during model training, thus addressing the mentioned key question in real-world traffic safety data collection:  
    
    % These advancements aim to improve data quality for researchers and offer valuable insights and tools for decision-makers within the broader traffic safety research community. 
    % This evidence-based guidance informs more strategic data collection and record management, improving data quality and strengthening the overall crash prediction and analysis process through continual refinements to the \trafficsafe framework.
\end{itemize}

%% file: results.tex
\section{Results}

\subsection{Multi-modal Crash Data}
\label{sec:raw_data}
% Our dataset comprises crash data from Washington State in 2022, totaling 16,188 records, and from Illinois in 2022, totaling 42,715 records, after excluding cases with missing key attributes related to vehicle or crash object status. Primary sources include the \textbf{Highway Safety Information System (HSIS) crash data} \supercite{hsis}, \textbf{satellite images} \supercite{googlemaps}, \textbf{police crash reports} \supercite{police_crash_reports}, and \textbf{crashworthiness data} \supercite{crashworthiness_data}. The HSIS crash data contains two main components: one describing the physical layout of roads and associated traffic characteristics in Washington and Illinois, such as the road level and speed limit, and the other containing crash reports with general crash descriptions such as crash location and time. Satellite images are obtained based on the location information given in the HSIS data. By processing these satellite images, more detailed descriptions of the road infrastructure and planning information, such as intersection alignments and neighborhood types, can be obtained. Police crash reports contain detailed descriptions of crashes, such as injury level and so on. Crashworthiness data includes on-site crash information and vehicle details, such as defects. Details of raw data formats and types are provided in Section \ref{sec:methods:raw_data}.

Our cleaned dataset comprises crash data from Washington State in 2022, totaling 16,188 records, and from Illinois in 2022, totaling 42,715 records, after excluding cases with missing key attributes related to vehicle or crash object status. Primary sources include the \textbf{Highway Safety Information System (HSIS) crash data} \supercite{hsis} and \textbf{satellite images} \supercite{googlemaps}. The HSIS crash data contains four major components: crash data, infrastructure data, vehicle data, and the person data. Crash data provides detailed descriptions of crashes, such as location, time, and injury severity. Infrastructure data includes information about road layouts and traffic characteristics, such as road level and speed limits. Vehicle data contains details such as manufacturing year and reported defects of the involved vehicles, while person data captures demographic and other relevant details about drivers and passengers, such as age and gender. Satellite images complement the HSIS data by providing additional visual context, including information about lanes, intersections, and other roadway attributes. Further information on raw data formats and types is available in Section \ref{sec:methods:raw_data}.

% , the \textbf{police crash reports} \supercite{police_crash_reports}, \textbf{crashworthiness data} \supercite{crashworthiness_data}, vehicle dynamic data\supercite{police_crash_reports}

\subsection{TrafficSafe Crash Outcomes Prediction Pipeline}

\begin{figure}[h!]
    \centering
    \includegraphics[width=1\linewidth]{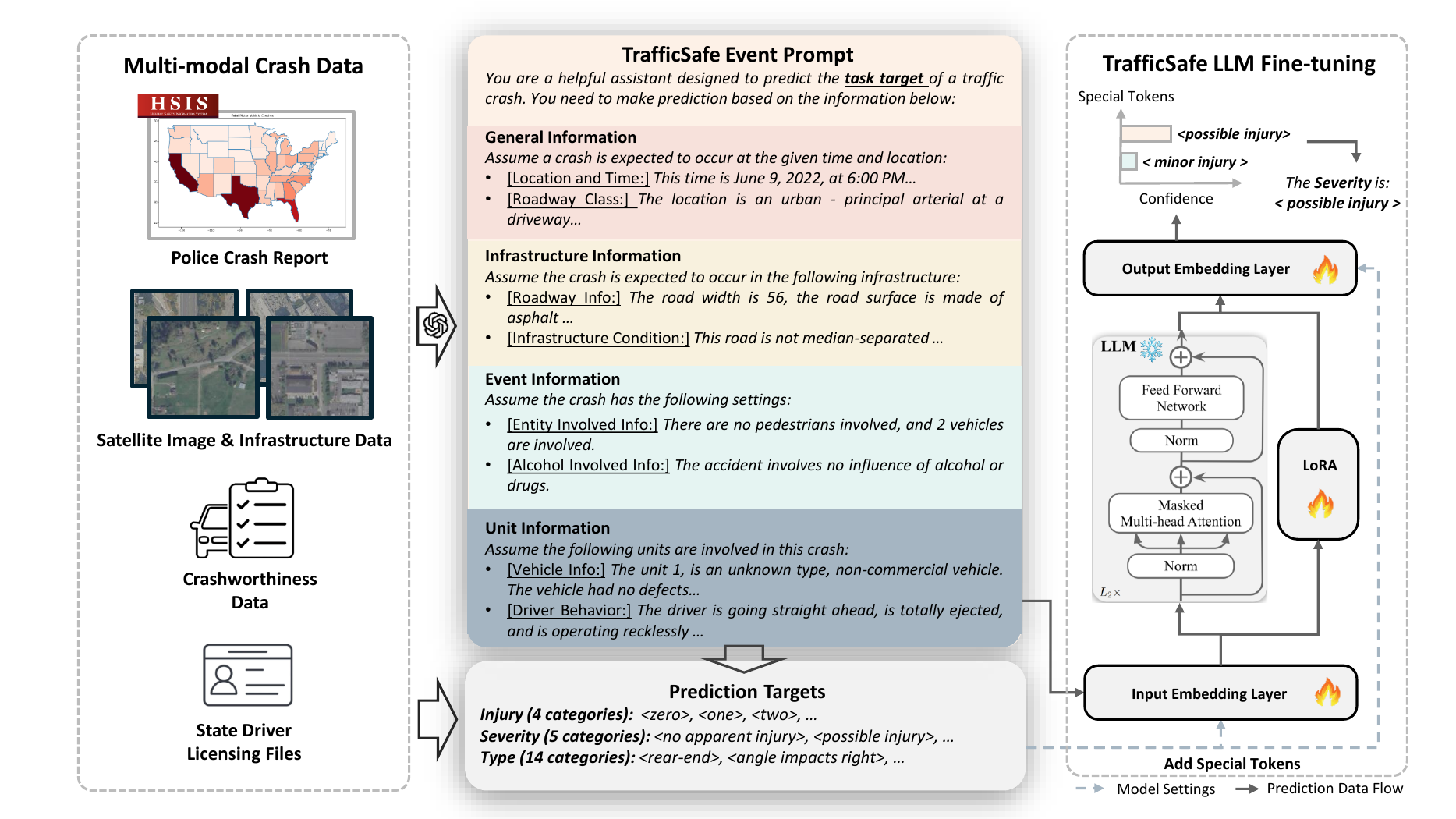}
    \caption{\textbf{\trafficsafe Crash Outcomes Prediction Pipeline.} Multi-modal crash data is collected and organized into textual prompts through an AI-expert cooperative process. The HSIS crash data, satellite images, and infrastructure data are used to extract general and infrastructure information, including the crash time, location, the road level, and so on. The vehicle data, and person data are converted into the event information and the unit information, including vehicle movements, driver characteristics (e.g., age, gender, alcohol use), vehicle attributes (e.g., manufacture year), and so on. \dataset is created with three prediction targets: \textit{Injury}, \textit{Severity}, and \textit{Type}. The \textit{Injury} task predicts the number of people injured in the crash event, the \textit{Severity} task estimates the severity level of the crash, such as \textit{no apparent injury} or \textit{fatal}, and the \textit{Type} task classifies type of crash, such as \textit{single vehicle with object} or \textit{angle impacts right} (The crash event consequences classification are provided in Supplementary Table 2 and Supplementary Table 3. The \model is fine-tuned using the \dataset. To reframe the crash outcomes prediction from a classification task to a language inference task, \model is fine-tuned by adding prediction targets as special tokens in its vocabulary and adjusting parameters using Low-Rank Adaptations (LoRA) \supercite{hu2021lora}. }
    \label{fig:2}
\end{figure}

To leverage the multi-modal crash data described in Section \ref{sec:raw_data} for crash prediction, we developed the \trafficsafe crash outcomes prediction pipeline, which transforms crash outcomes prediction into a text-based reasoning task. To achieve this, the raw crash data is organized into the textual \dataset, which is then used to fine-tune the \model. Figure \ref{fig:2} presents an overview of the \trafficsafe crash outcomes prediction pipeline, with subsequent sections detailing each stage of the pipeline.

% organize the \dataset and develop \model to learn and predict crash outcomes from textual prompts. 

% To leverage multi-modal crash data for crash prediction, we organize the \dataset and develop \model to learn and predict crash outcomes from textual prompts. 

\subsubsection{Constructing Prompts and Prediction Targets}
\label{sec:results:Constructing}
% The \dataset is created through an AI-expert cooperative textulization process, organizing multi-modal raw data for effective crash prediction. The detail information about the raw data and the textulization process are described in Section XXX. Finally, the raw data for each crash event case is organized as text content with four parts: 

The \dataset is created through an AI-expert cooperative textualization process, organizing multi-modal raw data for effective crash prediction. The detailed information about the raw data feature engineering and the textualization process are available in Section \ref{sec:methods:Feature_engineering}. As shown in Figure \ref{fig:2}, the constructed prompts are divided into five parts: one system prompt and four content parts, with each content part containing approximately 100 words. These parts include:

\begin{itemize}
    \item \textbf{System Prompt:} Provides an introduction and task-specific instructions. 
    \item \textbf{General Information:} Includes general information about the time and location of the prediction region and the roadway category.
    \item \textbf{Infrastructure Information:} Describes road infrastructure, encompassing static features like the number of lanes and speed limits, as well as dynamic elements such as work zones, lighting, and road surface conditions.
    \item \textbf{Event Information:} Contains detailed descriptions of crash events, such as the number of vehicles involved and their directions of movement.
    \item \textbf{Unit Information:} Provides vehicle and individual details relevant for crash prediction, such as airbag status and the driver's age.
\end{itemize}

The prediction targets consist of three variables: \textit{Injury}, \textit{Severity}, and \textit{Type} (see Figure \ref{fig:2}) \supercite{abdel2005analysis, IRANITALAB201727, SAVOLAINEN20111666}. Specifically, \textit{Injury} task predicts the number of people injured in the given crash event. \textit{Injury} task is treated as a classification task with four categories: \textit{zero}, \textit{one}, \textit{two}, and \textit{three or more than three}, where crashes involving more than two injured people are grouped into a single category due to the limited number of such cases. The \textit{Severity} task assesses the level of injury severity in a crash, classified into five levels from \textit{no apparent injury} to \textit{fatal}. \textit{Type} task predicts the type of crash, such as the \textit{rear-end collision} or \textit{collision with object}, with 14 crash type categories in the Washington dataset and 16 in the Illinois dataset. Detailed information on the defined targets is available in Section \ref{sec:methods:targets}. 

For each crash event, we perform the feature engineering and textualization process, organize the textualized data as input, and process labels corresponding to three tasks. The complete prompt examples are presented in Extended Data Figure \ref{fig:WA_prompt} and Extended Data Figure \ref{fig:IL_prompt}. Ultimately, after filtering out data items with missing information, the \dataset merges the complementary information from multi-modal data sources and contains 58,903 crash records with approximately 12.74 million words. These records are split into training, validation, and test sets in a 7:1.5:1.5 ratio. 

% The detailed dataset split process is shown in Section \ref{sec:data_split}. 

% \begin{itemize}
%     \item The number of people \textcolor{orange}{injured} $n_i^\mathcal{D} \in \{f(l)|l=0,1,2,\cdots\}$, where i denotes the $i$-th data in the dataset, $\mathcal{D} \in \{\mathcal{W}, \mathcal{I}\}$ denotes the Washington dataset $\mathcal{W}$ or the Illinois dataset $\mathcal{I}$, $l$ represents the number of people injured,  and $f(l)$ denotes the prediction targets when the injured people is $l$.
%     \item The \textcolor{blue}{severity} of the crash on the KABCO scale~\footnote{https://highways.dot.gov/media/20141} $s_i^\mathcal{D} \in \{ S_k|k=1,2,\cdots\ \}$, where $S_k$ is the $k$-th level of crash severity.
%     \item The \textcolor{purple}{crash type} $ t_i^\mathcal{D} \in \{ T^\mathcal{D}_k|k=1,2,,\cdots\ \}$, where $T^\mathcal{D}_k$ is the $k$-th label of crash type in dataset $\mathcal{D}$.
% \end{itemize}

% We utilize these three variables to describe the crash result \(\text{CR}_i^{\mathcal{D}}\). The crash outcome can be presented in the following format: $\text{CR}_i^{\mathcal{D}} = (n_i^\mathcal{D}, s_i^\mathcal{D},t_i^\mathcal{D})$. For numerical variables, the function \( f(l) \)  describes the number of people injured in crash as follows: zero if \(l=0\), one if \(l=1\), two if \(l=2\), and more than two if \(l \geq 3\), the values for $S_k$ and $T^\mathcal{D}_k$ are provided in the Section XXX.

% TrafficSafe leverage textualized multi-model crash data
% We use AI-expert cooperative method to textualize the multi-modal crash data. 

\subsubsection{Adapting LLM for Crash Prediction}

Although vanilla LLMs like Llama 3 possess broad general knowledge and strong text reasoning capabilities, they demonstrate limited effectiveness on crash prediction tasks without the fine-tuning process (see Supplementary Section 1). To address this, we developed \model, a specialized model fine-tuned on the processed \dataset. This fine-tuning process enhances the LLM's comprehension of crash events and enables accurate outcome prediction. Specifically, special tokens are introduced into the LLM vocabulary as prediction targets (Number of \textit{Injury}, \textit{Severity}, and Crash \textit{Type}), fine-tuning the model to generate these tokens during prediction. The details of the fine-tuning are provided in Section \ref{sec:methods:model}.

% \subsubsection{Experiment setup}

% We split the Washington and Illinois dataset into training, validation, and test set in a 7:1.5:1.5 ratio. For Washington state, 11,332 records are used for training, 2428 for validation, and 2428 for testing (Duo to data imbalance, 1428 records in validation and testing set and dropped, see Section XXX). For Illinois state, 29,307 records are used for training, 6,704 for validation, and 6,704 for testing. 

% \subsubsection{Evaluation of \model and reference models}

% \textbf{Evaluation metrics: } As we adapt the crash outcome prediction as a discriminative tasks, three common classification metrics are used to evaluate the performance of \model, which are accuracy, precision, and F1-score. The final metrics are calculated as the macro-averaged values across all categories. Recall is not included, as it yields the same value as accuracy under macro-averaging. The detail information of the metrics is documented in Section XXX.

% \textbf{Reference model: }

\subsection{Performance Evaluation of TrafficSafe LLM}
\label{sec:performance}

In this section, we evaluate the performance of \model and compare its performance with other baselines (see Section \ref{sec:methods:baselines}). The fine-tuning process is based on two vanilla LLMs with different sizes: Llama 3.1 8B and Llama 3.1 70B. Accuracy, precision, and F1-score are used as the evaluation metrics, the detail information is available in Section \ref{sec:methods:metrics}.

\input{tables/results}

\begin{figure}[p]
    \centering
    \begin{minipage}{.48\textwidth}
        \centering
        \includegraphics[width=\linewidth]{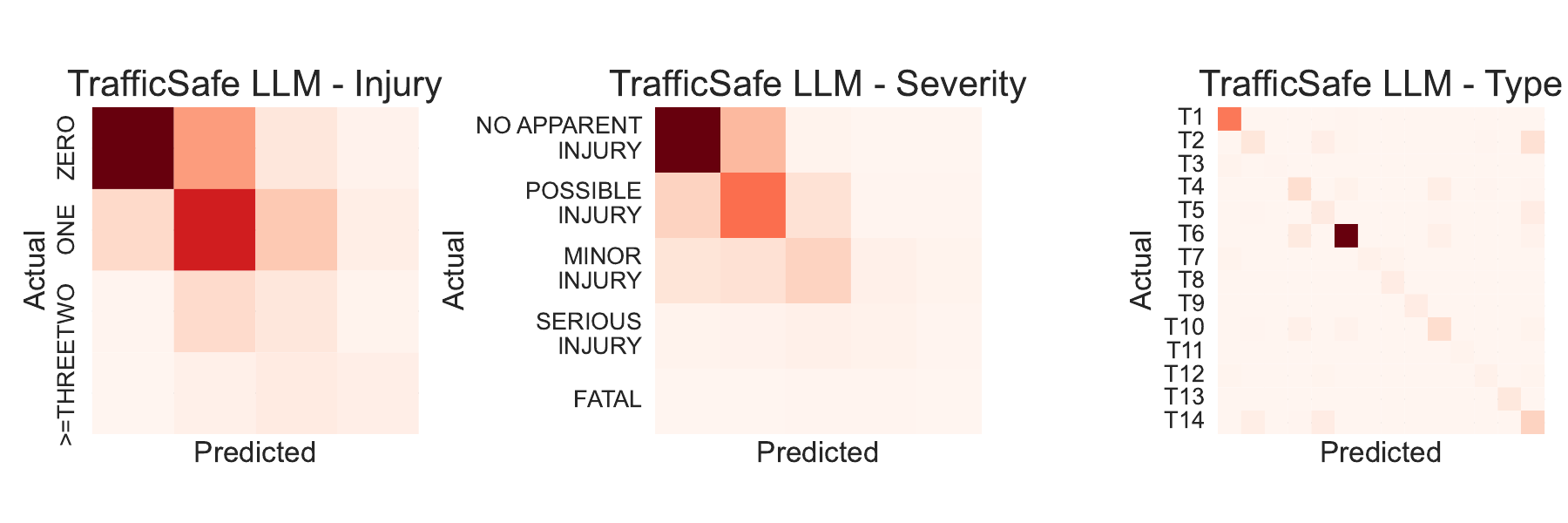}
        \subcaption{}
        \label{fig:cm_wa}
    \end{minipage}%
    \hspace{0.02\textwidth} % Adjust space between columns
    \begin{minipage}{.48\textwidth}
        \centering
        \includegraphics[width=\linewidth]{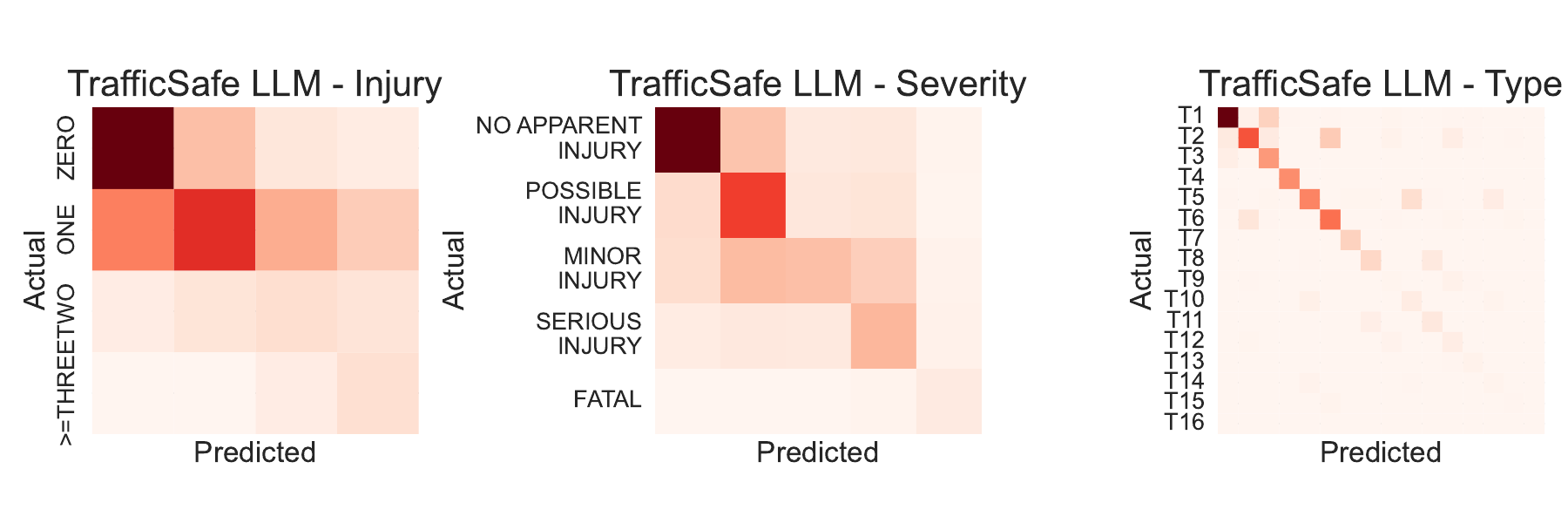}
        \subcaption{}
        \label{fig:cm_il}
    \end{minipage}
    
    % \vspace{0.5cm} % Adjust space between rows
    
    \begin{minipage}{.48\textwidth}
        \centering
        \includegraphics[width=\linewidth]{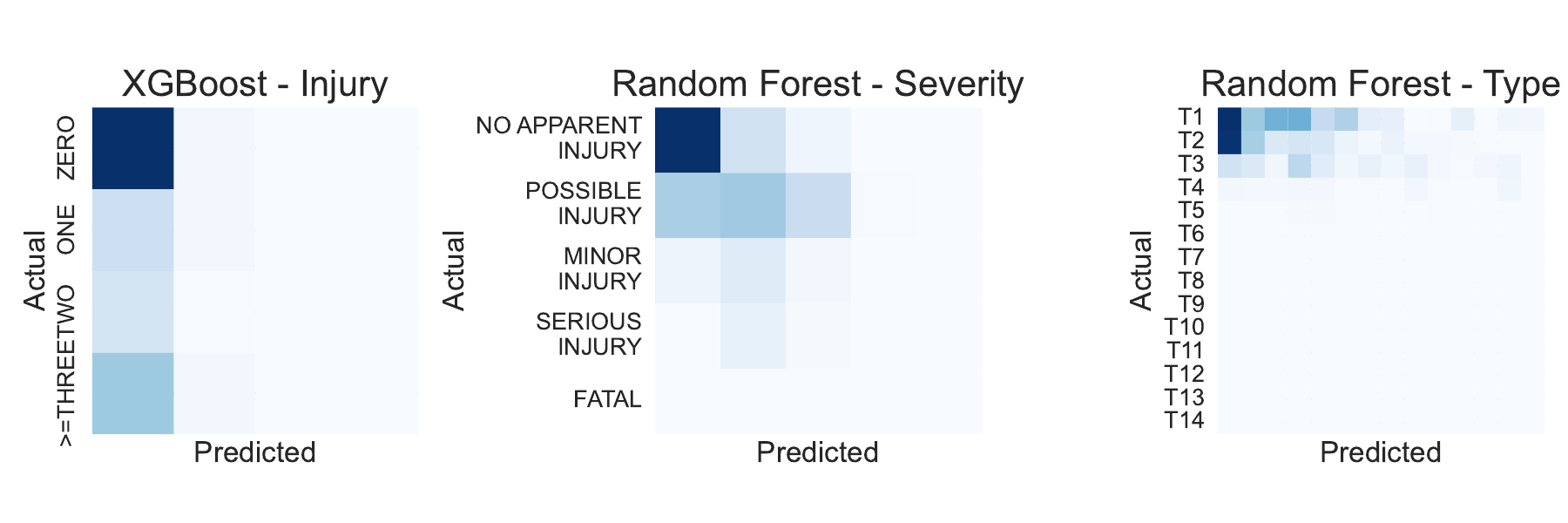}
        \subcaption{}
        \label{fig:cm_wa_b}
    \end{minipage}%
    \hspace{0.02\textwidth}
    \begin{minipage}{.48\textwidth}
        \centering
        \includegraphics[width=\linewidth]{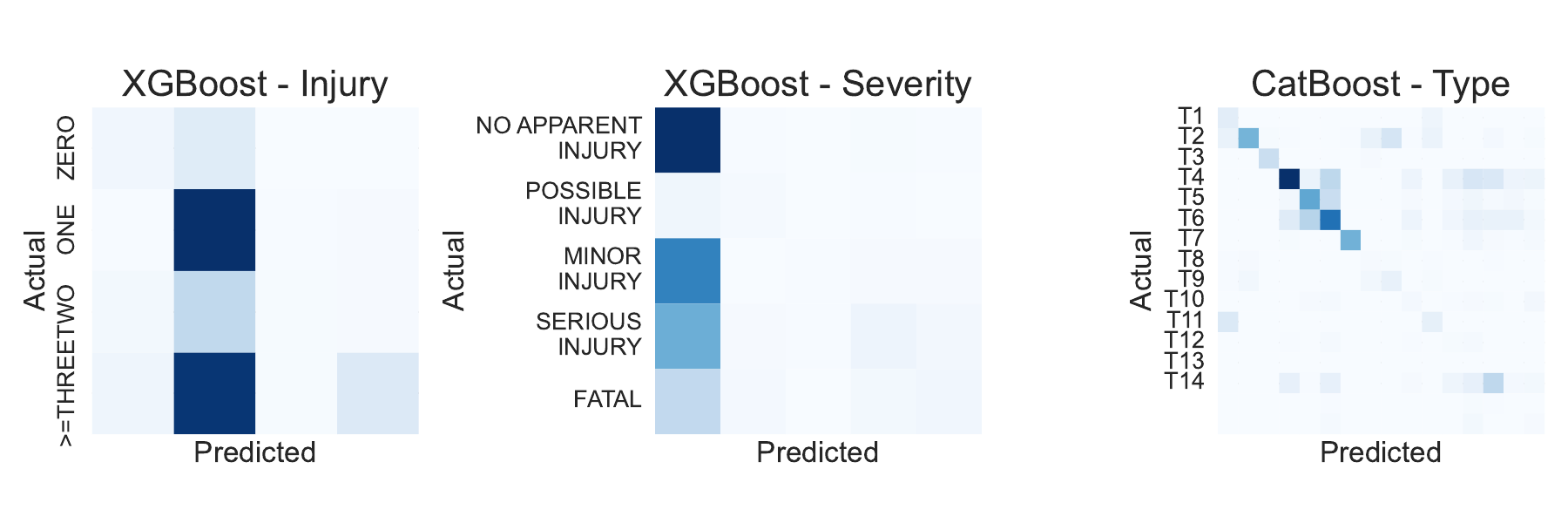}
        \subcaption{}
        \label{fig:cm_il_b}
    \end{minipage}

    \begin{minipage}{.48\textwidth}
        \centering
        \includegraphics[width=.9\linewidth]{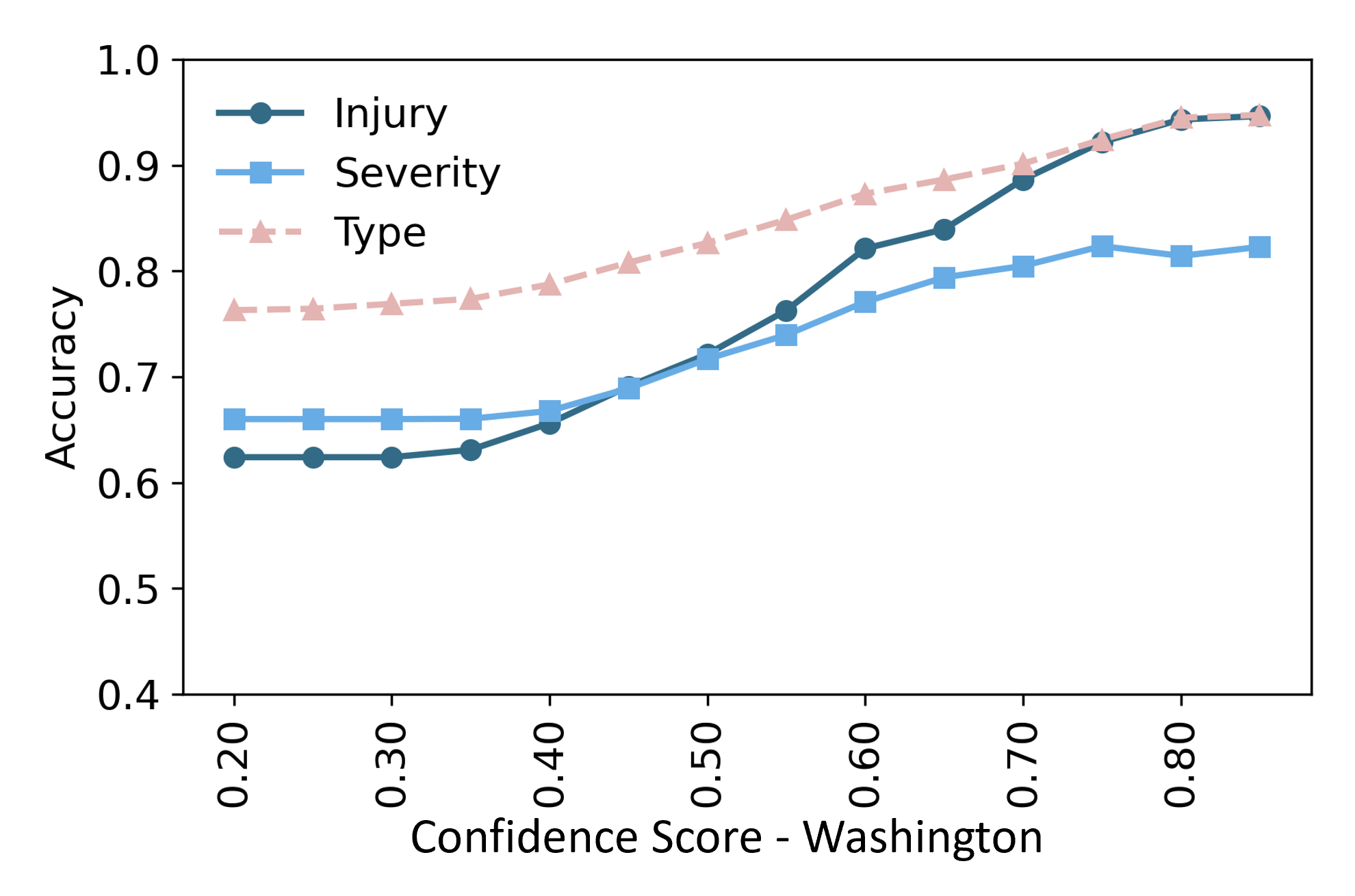}
        \subcaption{}
        \label{fig:cs_wa}
    \end{minipage}%
    \hspace{0.02\textwidth} % Adjust space between columns
    \begin{minipage}{.48\textwidth}
        \centering
        \includegraphics[width=.9\linewidth]{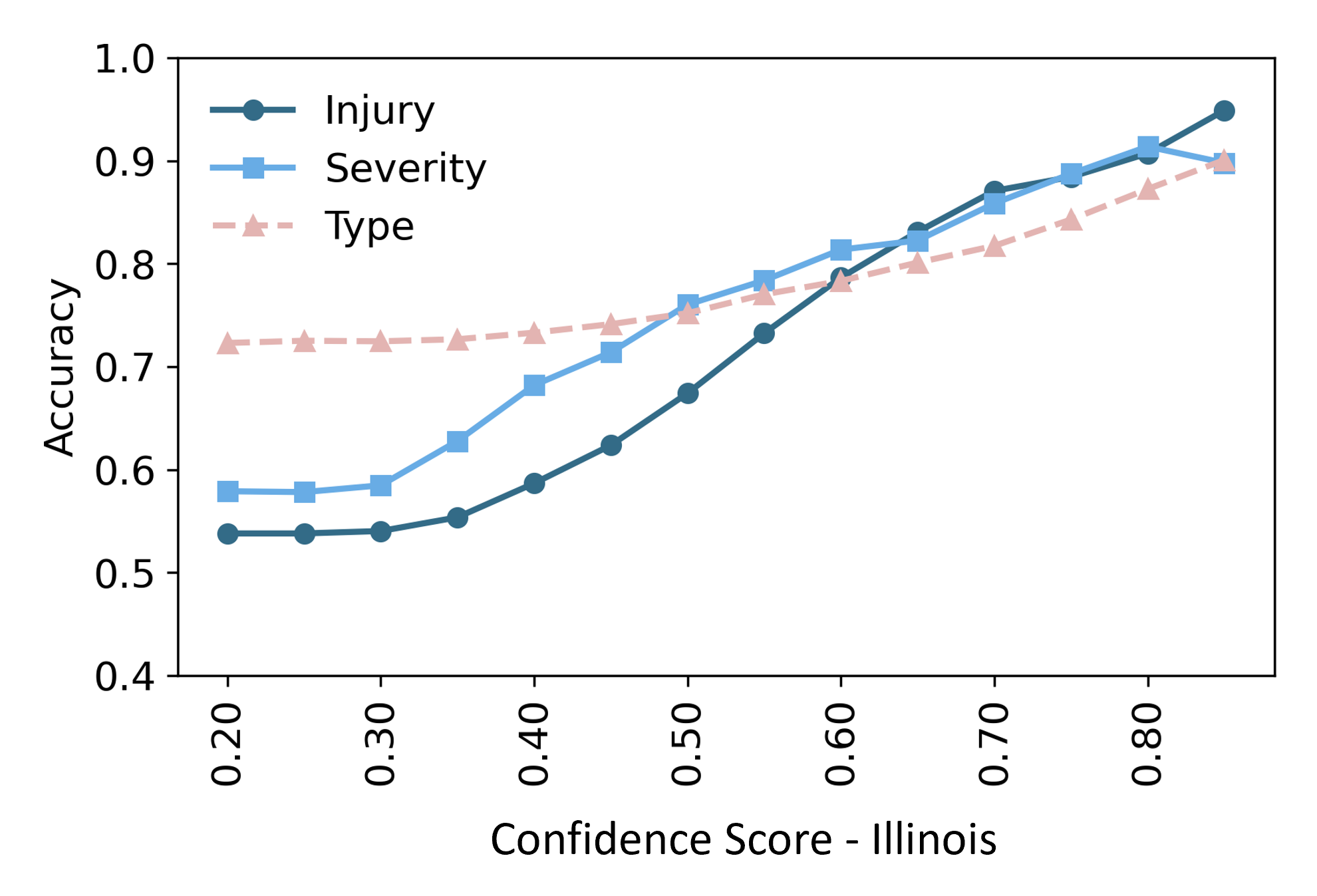}
        \subcaption{}
        \label{fig:cs_il}
    \end{minipage}
    \begin{minipage}{.32\textwidth}
        \centering
        \includegraphics[width=\linewidth]{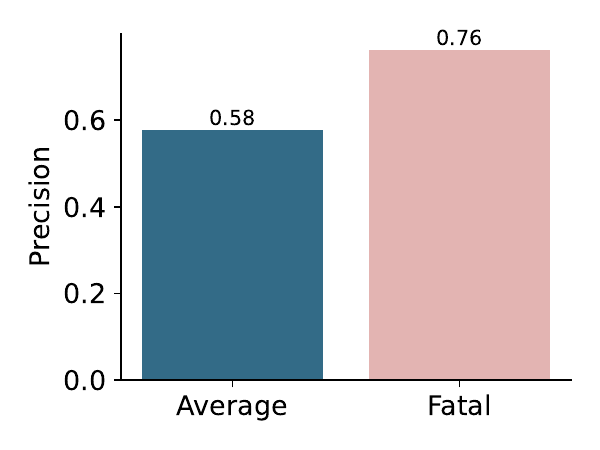}
        \subcaption{}
        \label{fig:fatal_il_acc}
    \end{minipage}
    % \hspace{0.02\textwidth} % Adjust space between columns
    \begin{minipage}{.32\textwidth}
        \centering
        \includegraphics[width=\linewidth]{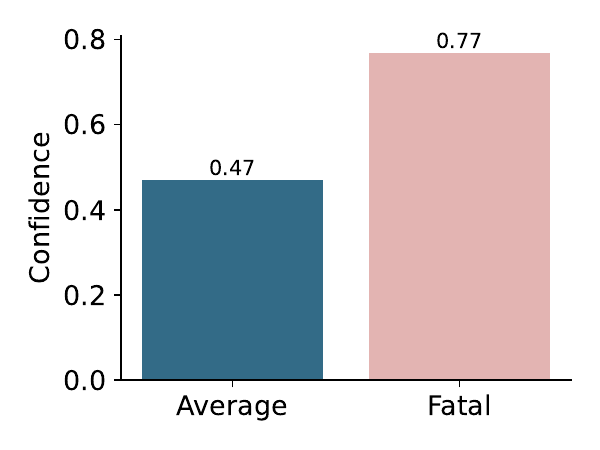}
        \subcaption{}
        \label{fig:fatal_il_conf}
    \end{minipage}
    \begin{minipage}{.32\textwidth}
        \centering
        \includegraphics[width=\linewidth]{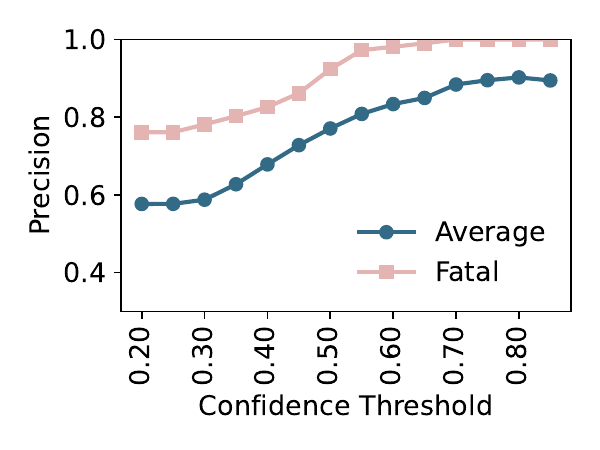}
        \subcaption{}
        \label{fig:fatal_il}
    \end{minipage}%

    \caption{\textbf{\model Provides Accurate and Trustworthy Predictions.} \model produces robust confusion matrices for both the (a) Washington and (b) Illinois datasets (we select the best results for each task by F1-score). In contrast, baseline models tend to predict the most frequent category across both the (c) Washington and (d) Illinois datasets (we show baseline models with the best F1-score. The performances for other baseline models can be found in Extended Data Figure \ref{fig:extended_cm}). Meanwhile, \model produces trustworthy predictions for both the (e) Washington and (f) Illinois datasets. Higher confidence levels in the model's predictions correspond to an increased likelihood of accuracy. Furthermore, (g) The \model achieves higher precision for fatal crash predictions. (h) Fatal crash predictions also exhibit higher confidence compared to average predictions in Illinois dataset. The Washington dataset is not shown due to limited fatal cases. (i) For fatal crashes, the \model achieves near-perfect precision (97.61\%) when the confidence score exceeds 0.6, indicating that the \model can deliver highly accurate and trustworthy predictions for fatal crashes.}
    \label{fig:grid}
\end{figure}

\textbf{\model provides accurate crash predictions, even in zero-shot scenarios.} Table \ref{tab:maintable} compares the performances of \model and adopted baselines. The results show that the \model outperforms all the baselines in each task setting with an average F1-score improvement of 41.7\% across multiple tasks. Specifically, in the crash \textit{Type} prediction task in Washington dataset, the \model achieves F1-score of 0.759, which is more than 130\% higher than all other comparative methods. \model performs well on both the Washington and Illinois datasets, demonstrating its stability across diverse geographical regions. Moreover, as shown in the confusion matrix in Figure \ref{fig:cm_wa} and Figure \ref{fig:cm_il}, beyond improved metrics, \model demonstrates a more balanced prediction distribution. In contrast, as shown in Figure \ref{fig:cm_wa_b} and Figure \ref{fig:cm_il_b}, traditional machine learning models tend to predict the dominant categories (e.g., \textit{zero} under \textit{Injury} prediction task, \textit{no apparent injury} under \textit{Severity} prediction task). The complete confusion matrix is shown in Extended Data Figure \ref{fig:extended_cm}. Moreover, the ability of a model to generalize across unseen scenarios is vital to ensuring its robustness and applicability in real-world contexts. We tested \model generalization ability by using the \model trained on Illinois dataset and evaluating its performance on the unseen Maine and North Carolina datasets. Notably, without additional fine-tuning, \model achieved F1-scores averaging 0.542 in North Carolina and 0.521 in Maine, closely matching its performance in Illinois (see Table \ref{tab:maintable}). This underscores \model's ability to generalize well to previously unseen datasets, further validating its potential for real-world applications.
% Furthermore, \model demonstrates its effectiveness in zero-shot scenarios. As shown in Table \ref{tab:maintable}, \model outperforms baseline models on North Carolina dataset and Maine dataset in F1-score, which were not included during the training phase. This underscores \model's ability to generalize well to previously unseen datasets, further validating its potential for real-world applications.

% \model delivers accurate predictions, forming a solid foundation for further analysis.

 % We observe that increasing model size enhances performance, with the 70B model achieving the highest rank across both Washington and Illinois datasets.

% \begin{figure}
%     \centering
%     \includegraphics[width=1\linewidth]{imgs/confusion.png}
%     \caption{\Yang{@Yibo: Four figures, larger font, remove numbers, order: Injury, severity, type}}
%     \label{fig:confusion}
% \end{figure}

% \begin{figure}
%     \centering
% \end{figure}

\textbf{\model provides trustworthy crash predictions, where a higher confidence score links to higher accuracy.} \model tailors LLMs for discriminative crash outcomes prediction tasks, generating predictions accompanied by confidence scores that represent the probabilities associated with specific special tokens. Figure \ref{fig:cs_wa} and Figure \ref{fig:cs_il} illustrate the trend of accuracy in relation to the confidence scores of \model's predictions for the Washington and Illinois datasets. The results indicate that our model achieves greater accuracy at higher confidence levels. For instance, for the \textit{Injury} prediction task in the Washington dataset, when the model’s confidence score exceeds 0.40, the accuracy rises above 0.65, and with confidence scores over 0.60, the accuracy surpasses 0.80. The strong positive correlation between confidence scores and accuracy showcases the trustworthiness of the \trafficsafe framework. By providing reliable confidence scores alongside predictions, the framework empowers informed decision-making in real-world applications.

% This feature demonstrates the \trafficsafe framework's capability to provide trustworthy predictions and facilitate decision-making by offering confidence scores that closely align with the correctness of crash event predictions, making it highly suitable for real-world applications.

% # 定义confidece score，加一句acc与confidence score的关系

% \subsection{Data contribution at training phase}

% \Yang{@Pu: person \& unit data or unit info, consistent with the content before}

% table~\ref{tab:train_shapley}. The results indicate that, for the severity task, the person \& unit data detailing attributes of the primary entity involved in the crash have the highest contribution to the model's performance, followed by the event data, which provides information on the vehicle's movement prior to the accident. For the acctype task, the event data directly related to vehicle movement has the highest contribution, followed by the person \& unit data.

% \subsection{Crash factor attribution}
\subsection{TrafficSafe Attribution and Result Interpretation}
\label{sec:attribution}

\begin{figure}[pbth]
    \centering
    \begin{subfigure}[b]{1\textwidth}
        \centering
        \includegraphics[width=0.88\textwidth]{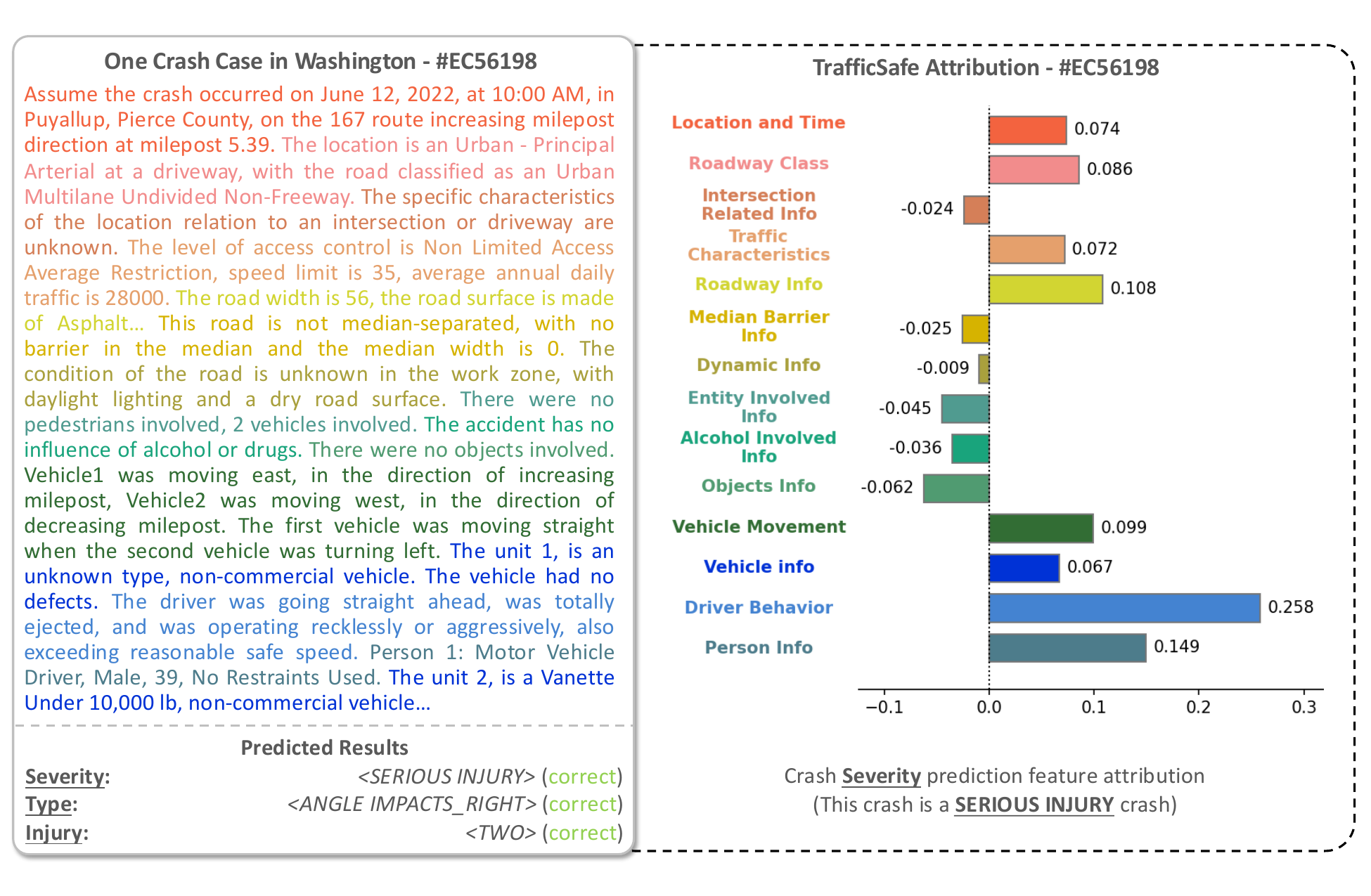}
        \vspace{-0.3cm}
        \caption{\textbf{Sentence-based Feature Attribution Results for a Crash Resulting in Serious Injuries in Washington Dataset.}}
        \label{fig:WA_shap}
    \end{subfigure}

    \begin{subfigure}[b]{1\textwidth}
        \centering
        \includegraphics[width=0.88\textwidth]{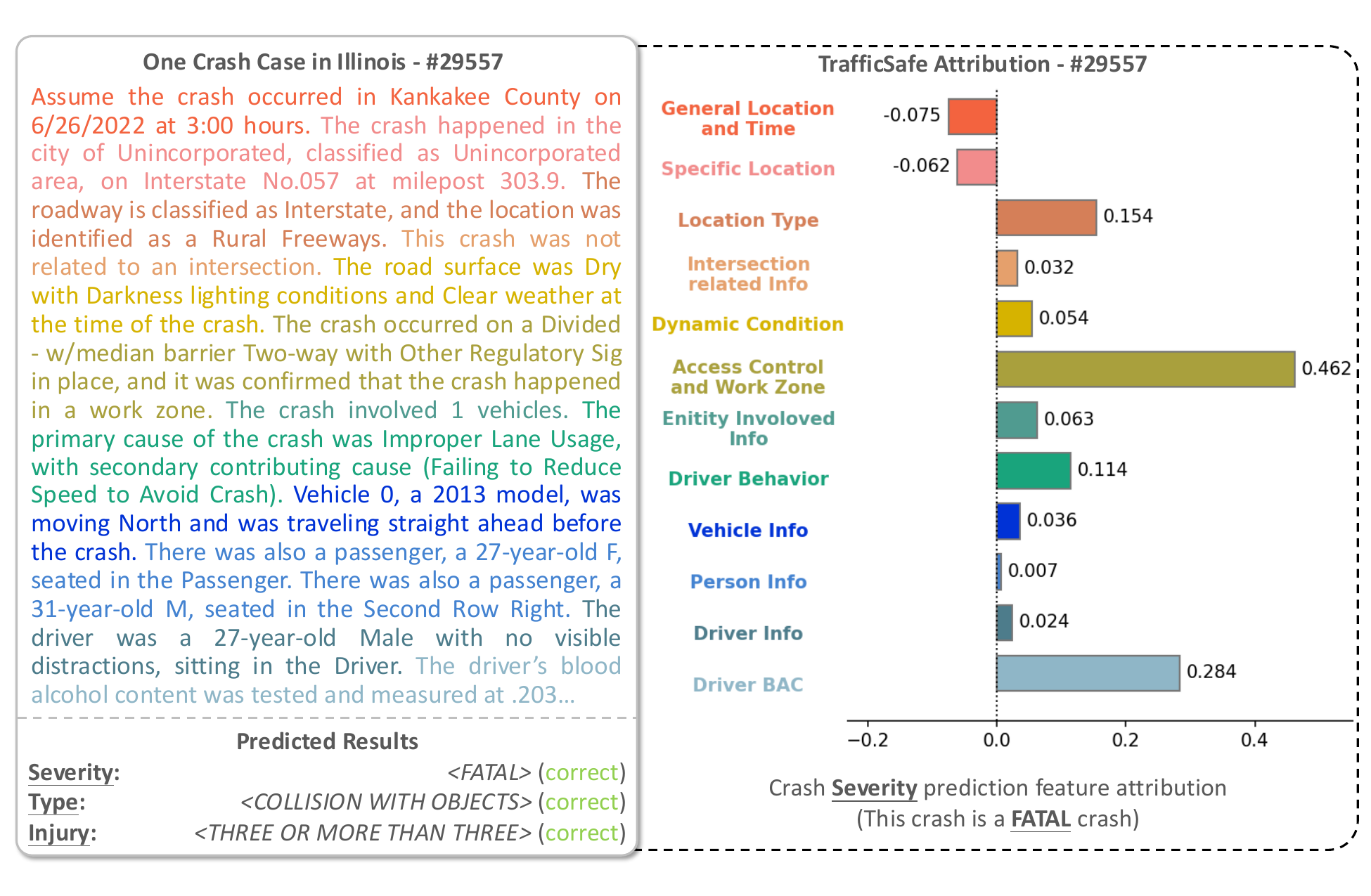}
        \vspace{-0.2cm}
        \caption{\textbf{Sentence-based Feature Attribution Results for a Crash Resulting in Fatalities in Illinois Dataset.}}
        \label{fig:IL_shap}
    \end{subfigure}

    \caption{\textbf{Single Case Feature Attribution Results for \textit{Severity} Task.} The left part displays the full prompt from (a) Washington and (b) Illinois, with different colors representing various semantic text sequences. The right part illustrates the feature contribution assigned to each text sequence. Positive contributions signify a supportive role in the model's prediction, whereas negative contributions indicate a detracting influence. The absolute value of these contributions represents the importance of each sequence to the model's output. }
    \label{fig:Severity_single_shap}
\end{figure}

Understanding how \model generates accurate predictions and how various components of the input prompt influence the outcomes is fundamental to enabling evidence-based decision-making. As shown in figure~\ref{fig:cs_wa} and~\ref{fig:cs_il}, the \model’s confidence score strongly correlates with its predictive accuracy for fatal and serious injury crashes, therefore, we can use the confidence score to represent a case’s real-world risk level. Notably, the \model’s confidence scores tend to be lower than their corresponding precision values (see figure ~\ref{fig:cs_wa} and~\ref{fig:cs_il}, indicating that the confidence score is a conservative estimate of risk). 

Within the \attribution framework, a sentence-based feature contributions calculation method was proposed to identify how each sentence contributes to the LLM’s outputs based on Shapley theory which is recognized as a systematic and equitable method for attributing the contribution of each feature to a model’s output \supercite{bordt2023shapley, shapley:book1952}, thereby revealing crash-related factors at the event level (see Section \ref{sec:method:attribution} for details). In essence, each feature's contribution represents its share of responsibility for the model's confidence in a particular prediction. The sum of all feature contributions equals the confidence score itself. Figure \ref{fig:Severity_single_shap} illustrates sentence-level feature contributions for the severity of individual crash events, using one crash from Washington and one from Illinois as examples. In the Washington crash example (Figure~\ref{fig:WA_shap}), \textit{Driver Behavior} (e.g., reckless driving or speeding) is the primary factor contributing to serious injury crashes with the feature contribution of 0.258. \textit{Person Info} (e.g., no seatbelt use) also shows a substantial impact with the feature contribution of 0.149. By contrast, \textit{Dynamic Info} (daylight and dry roads) lowers the probability of crash with serious injuries with a negative feature contribution of -0.009. While, in the Illinois example (Figure~\ref{fig:IL_shap}), an elevated \textit{BAC} (Blood Alcohol Content, with feature contribution of 0.284) and the presence of a \textit{Work Zone} (feature contribution of 0.462) notably increase the likelihood of fatal crash outcomes. Beyond the above examples, more additional sentence-level feature attribution analysis can be found in Extended Data Figure \ref{fig:WA_FA2}, \ref{fig:IL_FA1}, Supplementary Section 5 and 6.

%given the strong correlation between the \model's confidence score and its predictive accuracy for fatal and serious injury crashes, we use this score to represent a case’s real-world risk level. Actually, the \model confidence score are always much lower compared to the precision, referring a comprised risky level, 60\% confidence score leading to the over 95\% precision. As a result, the degree to which a specific factor influences the prediction can be directly interpreted as its real-world risk contribution. 

% This sentence-level feature contributions calculation method provides a novel approach for conducting combined text-based contribution analysis and supports more informed decision-making in traffic safety. 

The following sections utilize \attribution framework to examine feature importance from two perspectives: 1) at the inference stage, to identify key factors influencing crash predictions under various conditions and high-risk scenarios, and 2) at the training stage, to understand which data components are most important for model learning.

\subsubsection{Factor Attribution at Inference Stage for Conditional Risk Analysis}

\begin{figure}[p]
    \centering
        \centering
        \includegraphics[width=\textwidth]{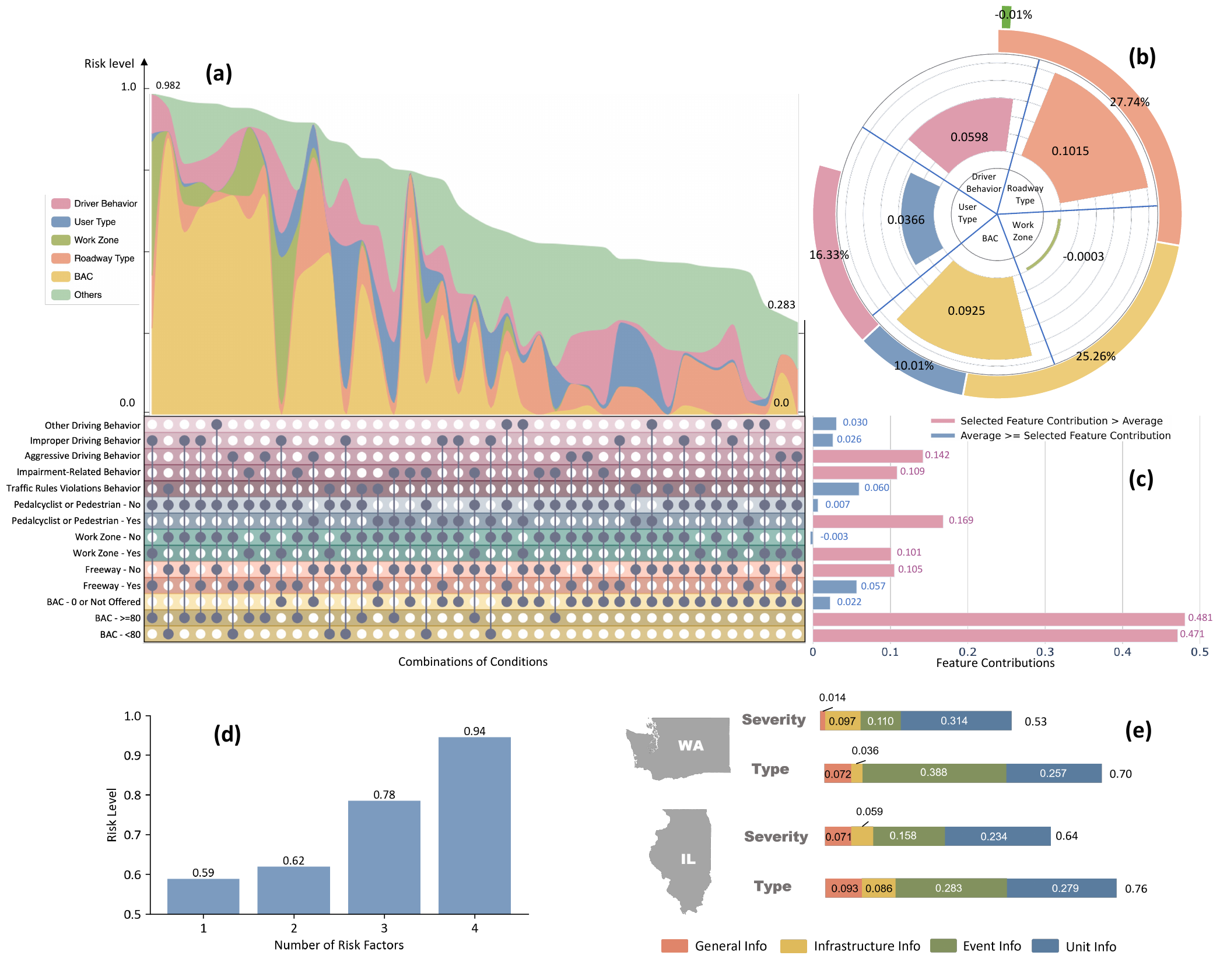}

        \vspace{-10pt}
    
    \caption{\textbf{Conditional Risk Analysis for the Serious Injury and Fatal Crashes.} Higher confidence scores in \model's predictions correspond to greater accuracy, allowing the confidence score (calculated as the sum of feature contributions for all data components) to serve as an indicator of risk level for serious and fatal crashes. (a) The estimated risk levels for various feature combinations are presented, with each level corresponding to the average confidence score of \model's predictions under the same conditions. Each column represents a specific combination of conditions (marked by dark dots) alongside the corresponding feature contribution for selected factors. (b) Feature contributions of five key factors and their proportions relative to all factors are visualized. The inner circle represents the average feature contribution of each factor across different values, while the outer circle shows the percentage share of each factor in the total average feature contribution. The unit for BAC (Blood Alcohol Content) is "mg/L", which is omitted in this figure.  %Some combinations are not shown due to insufficient data for visualization. 
    (c) Average feature contribution for each factor under specific values. Bars are marked in pink if the value exceeds the corresponding factor's average shown in (a) and in blue if it does not. (d) The strong correlation between the number of risk factors and the risk level of crashes. The high-risk factors are defined as driving after drinking (both BAC <= 80 mg/L and BAC > 80 mg/L), driving in work zones, driving on freeways, pedestrian-involved crashes, and high-risk driver behaviors (aggressive or impairment-related). For each case, we tallied the number of these risk factors and calculated the average risk level for all cases sharing the same count. (e) Feature contributions of different data components during the training stage for Washington and Illinois datasets.} 
    \label{fig:5}
\end{figure}

Conditional analysis evaluates crash outcomes across various scenarios, such as driving with or without alcohol consumption, to quantify the risk factors associated with each scenario. Severe crashes (serious injuries and fatal crashes) were prioritized in the conditional analysis due to their critical importance for traffic safety. These crashes, particularly fatal ones, were predicted accurately and reliably by \model\ (see Figures \ref{fig:fatal_il_acc}, \ref{fig:fatal_il_conf}, and \ref{fig:fatal_il}). 
% To ensure consistency, the Washington dataset was adapted to align with the Illinois prompt, resulting in 534 correctly predicted samples for analysis. 
Five key contributing factors were identified for this conditional analysis: \textit{Driver BAC} (BAC = 0 or not offered / BAC < 80 / BAC >= 80), \textit{Roadway Type} (Highway / not highway), \textit{Work Zone} (Work zone / not work zone), \textit{User Type} (Pedalcyclist or pedestrian / not pedalcyclist or pedestrian), and \textit{Driver Behavior} (Aggressive driving / impairment-related behavior / traffic rules violations / improper driving / others). Collectively, these factors accounted for an average of 79.33\% of the model’s overall attribution in predicting serious and fatal crashes (see Figure \ref{fig:5}b). A summary of key findings is provided:

\begin{itemize}
    \item \textbf{The BAC record emerges as a critical determinant in predicting serious and fatal crashes.} Among all contributing factors, BAC accounts for 25.26\% of the total contribution to serious and fatal crash prediction (see Figure \ref{fig:5}b). Notably, its contribution substantially increases when a driver consumes alcohol, irrespective of the amount. When drivers are under the influence of alcohol even if their BAC does not exceed the legal intoxication limit of 80 mg/L \supercite{washington_bac_limit,illinois_bac_limit}, this factor’s feature contribution still reaches approximately 0.45, surpassing that of most other factors in many cases (see Figure \ref{fig:5}a). Conversely, when a driver’s BAC is recorded as "zero or not offered," its contribution approaches zero, indicating minimal impact on the model's predictions.

    \item \textbf{Driving in a work zone is already risky under sober conditions, but alcohol consumption significantly increases the danger, making it one of the most hazardous scenarios for severe-injury crashes.} As shown in Figure \ref{fig:5}a, driving in a work zone while sober (“Work Zone-Yes” and “BAC = 0 or not offered”) contributes little to severe crash outcomes, with an average feature contribution of 0.03. However, after consuming alcohol (whether "BAC >= 80" or "BAC < 80"), the work zone feature contribution rises more than seven time to an average of 0.22. Furthermore, the overall crash risk increases substantially when driving in a work zone after drinking, as indicated by an average risk level of 0.78, compared to 0.44 under sober conditions. These findings indicate that work zones become especially hazardous when alcohol consumption is involved, creating one of the highest-risk scenarios for severe crash outcomes. Potential drunk driving warnings and risk mitigation strategies shall be closely linked with work-zone areas.
    
    % \item \textbf{Driving in a work zone under sober conditions presents a low level of risk, but consuming alcohol significantly amplifies this risk.}
    % Driving in a work zone is risky, and driving in a work zone after consuming alcohol is significantly much more risky compared to sober driving.} 
    % As shown in Figure \ref{fig:5}a, driving in a work zone after consuming alcohol (whether "BAC>=80" or "BAC<80") is associated with markedly higher confidence scores for predicting serious and fatal crashes compared to sober driving in the same conditions ("Work Zone-Yes" and "BAC-0 or not offered"). Specifically, the average feature contribution for driving in a work zone rises to 0.22 under the influence of alcohol, compared to 0.03 when driving sober. Similarly, the overall confidence score increases from 0.45 to 0.78, highlighting the heightened risk associated with alcohol consumption in work zone areas.
    % the average confidence score for driving under the influence of alcohol in work zones is 0.78, which is significantly higher than 0.45 for sober driving in work-zone areas. 
    % These results show that driving in a work zone after drinking constitutes one of the most risk scenarios for severe crash outcomes.
    
    \item \textbf{Aggressive driving and impairment-related behavior exhibit the highest contributions among driver behaviors.} Furthermore, combined with other conditions, aggressive and impairment-related behaviors pose nearly twice the risk for severe crash outcomes compared to other driver behaviors. As illustrated in Figure \ref{fig:5}c, aggressive driving emerges as the most significant contributor between driver behaviors, with feature contribution of 0.14. Impairment-related behavior, including driving under the influence of alcohol or drugs, also has a substantial influence, with average feature contribution of 0.11. In comparison, other improper driver behaviors, such as traffic rule violations (feature contribution of 0.07) and distractions like mobile phone use (categorized under improper driving, with feature contribution of 0.03), show below-average contributions to serious and fatal crashes. The "other" category, which includes normal driving and unknown behaviors, has the smallest impact, with feature contribution of 0.03.

    \item \textbf{The co-occurrence of risk factors significantly increases the expected crash risk level.} As illustrated in Figure \ref{fig:5}d, our analysis reveals a strong correlation between the number of risk factors present in a crash and the expected risk level for severe crash outcomes. When only one risk factor is involved, the risk level for severe crash outcomes is estimated at 0.59. This value increases to 0.62 with two risk factors, surges to 0.78 with three, and escalates to 0.94 when four risk factors co-occur. Notably, scenarios with three or more risk factors are markedly more dangerous than those with one or two. For example, while a combination of a BAC exceeding 80 mg/L and driving in a non-freeway work zone yields an average risk level of 0.51, substituting the non-freeway work zone with a freeway work zone increases the average risk level dramatically to 0.97. Such elevated risks indicate that the synergy among multiple factors is far from merely additive; instead, they appear to compound one another, amplifying the potential for severe outcomes. These findings show that transportation agencies need to prioritize multi-faceted interventions tailored specifically to scenarios with overlapping high-risk conditions.
    
    %These findings underscore the need for stricter traffic management measures in co-occurrence of high-risk factors, such as work zones and freeways, or work zones and driving with alcohol impact to mitigate severe crash outcomes.
    
    % These findings underscore the strong relationship between improper driver behavior and the likelihood of serious and fatal crashes, with aggressive driving and impairment-related behavior posing the greatest risks. 
\end{itemize}

\subsubsection{Factor Attribution at Training Stage for Effective Data Collection and Model Development}
\label{sec:causal_training}

% \textbf{Raw data formats of crash records may impact the attribution process.} The driver’s alcohol usage is a major contributing factor to crashes \supercite{report}. In the raw data records, both Washington and Illinois datasets include information on driver alcohol use. However, while the Washington dataset records alcohol use as a simple 'yes' or 'no' indicator, the Illinois dataset provides BAC as a numerical value. This variation in data format significantly affects the process of identifying attribution. Figure XXX shows sentence-level Shapley value distribution for \textit{Severity} prediction task across Washington and Illinois dataset. We observe that in the Washington dataset, alcohol usage does not emerge as a primary leading factor for any crash severity level. Conversely, in the Illinois dataset, alcohol usage stands out as the leading factor for fatal crashes, aligning more closely with real-world conditions. 

\textbf{Event information and unit information are the most important components for the model training.} While feature contributions at the inference stage reveal which features drive critical crash outcomes, understanding feature contributions during training provides deeper insights into which data components most effectively enhance model accuracy. As shown in Figure \ref{fig:5}e, the feature contributions of each component in the Washington and Illinois datasets are shown, demonstrating their impact on the model’s performance during training (see Supplementary Table 7 for detailed results and Section \ref{sec:method:attribution} for calculation details). The results indicate that in both the Washington and Illinois datasets, for the \textit{Severity} task, the unit information describing attributes of the primary entities involved in the crash has the highest contribution to the model's performance (0.314 in Washington, 0.234 in Illinois). Event information, which provides information on the vehicle's movement prior to the crash, is followed by unit information and has the second highest contribution (0.110 in Washington, 0.158 in Illinois). For the Crash \textit{Type} Prediction task, the event information has the highest contribution (0.388 in Washington, 0.283 in Illinois), followed by the unit information (0.257 in Washington, 0.279 in Illinois) and other components. These results can provide preliminary guidance on prioritizing the information collection for crash events, thereby improving crash prediction and feature attributions for better safety decision support.

%% file: tables/results.tex
\begin{table}[h]

    \centering
        \caption{\textbf{Performance Comparison of the three Crash Prediction Tasks on Washington Dataset and Illinois Dataset.} We present quality metrics along with model rankings by averaging the column-wise rank. The zero-shot results for North Carolina and Maine were derived using the model trained on the Illinois dataset. In supervised fine-tuning experiments on the Washington and Illinois datasets, \model outperforms all other methods, with \trafficsafe 70B achieving the best performance. Additionally, \model demonstrates strong generalization capabilities in zero-shot experiments on the North Carolina and Maine datasets.}
    \begin{subtable}{\textwidth}
        \centering
          % \caption{Washington dataset}
\label{tab:performance_WA}
\resizebox{1\textwidth}{!}{
\begin{tabular}{c|c|ccc|ccc|ccc|c}
\toprule[1.5pt]
\multirow{3}{*}{\textbf{Dataset}} & \multirow{3}{*}{\textbf{Model}} & \multicolumn{3}{c|}{\textbf{Injury}}              & \multicolumn{3}{c|}{\textbf{Severity}}             & \multicolumn{3}{c|}{\textbf{Type}}              & \multirow{3}{*}{\textbf{Rank}} \\ \cline{3-11}
                &       & \multicolumn{1}{c}{\textbf{Accuracy }} & \multicolumn{1}{c}{\textbf{Precision }} & \multicolumn{1}{c|}{\textbf{F1-score }} & \multicolumn{1}{c}{\textbf{Accuracy }} & \multicolumn{1}{c}{\textbf{Precision }} & \multicolumn{1}{c|}{\textbf{F1-score }} & \multicolumn{1}{c}{\textbf{Accuracy }} & \multicolumn{1}{c}{\textbf{Precision }} & \multicolumn{1}{c|}{\textbf{F1-score }} &                       \\ 
\midrule[1.5pt]
% \multicolumn{11}{c}{\textbf{Washington Dataset}} \\
% \midrule
\multirow{8}{*}{\textbf{Washington}} & RandomForest \supercite{randomforest}           & \multicolumn{1}{c}{0.522}              & \multicolumn{1}{c}{0.649}              & \multicolumn{1}{c|}{0.545}              & \multicolumn{1}{c}{0.628}              & \multicolumn{1}{c}{0.546}              & \multicolumn{1}{c|}{0.549}              & \multicolumn{1}{c}{0.740}              & \multicolumn{1}{c}{0.398}              & \multicolumn{1}{c|}{0.274}              & 4 (4.11)                \\
& AdaBoost \supercite{FREUND1997119}             & \multicolumn{1}{c}{0.495}              & \multicolumn{1}{c}{0.245}              & \multicolumn{1}{c|}{0.328}              & \multicolumn{1}{c}{0.492}              & \multicolumn{1}{c}{0.245}              & \multicolumn{1}{c|}{0.328}              & \multicolumn{1}{c}{0.563}              & \multicolumn{1}{c}{0.249}              & \multicolumn{1}{c|}{0.302}              & 8 (6.00)                \\
& CatBoost \supercite{prokhorenkova2019catboost}              & \multicolumn{1}{c}{0.495}              & \multicolumn{1}{c}{0.245}              & \multicolumn{1}{c|}{0.328}              & \multicolumn{1}{c}{0.492}              & \multicolumn{1}{c}{0.245}              & \multicolumn{1}{c|}{0.328}              & \multicolumn{1}{c}{0.715}              & \multicolumn{1}{c}{0.400}              & \multicolumn{1}{c|}{0.329}              & 6 (5.22)                \\
& DecisionTree \supercite{quinlan1986induction}          & \multicolumn{1}{c}{0.495}              & \multicolumn{1}{c}{0.245}              & \multicolumn{1}{c|}{0.328}              & \multicolumn{1}{c}{0.528}              & \multicolumn{1}{c}{0.428}              & \multicolumn{1}{c|}{0.372}              & \multicolumn{1}{c}{0.628}              & \multicolumn{1}{c}{0.406}              & \multicolumn{1}{c|}{0.323}              & 5 (4.67)                \\
& LogisticRegression \supercite{cox1958regression}   & \multicolumn{1}{c}{0.495}              & \multicolumn{1}{c}{0.245}              & \multicolumn{1}{c|}{0.328}              & \multicolumn{1}{c}{0.492}              & \multicolumn{1}{c}{0.245}              & \multicolumn{1}{c|}{0.328}              & \multicolumn{1}{c}{0.547}              & \multicolumn{1}{c}{0.401}              & \multicolumn{1}{c|}{0.309}              & 7 (5.67)                \\
& XGBoost \supercite{10.1145/2939672.2939785}               & \multicolumn{1}{c}{0.566}              & \multicolumn{1}{c}{0.665}              & \multicolumn{1}{c|}{0.469}              & \multicolumn{1}{c}{0.534}              & \multicolumn{1}{c}{0.428}              & \multicolumn{1}{c|}{0.367}              & \multicolumn{1}{c}{0.739}              & \multicolumn{1}{c}{0.413}              & \multicolumn{1}{c|}{0.298}              & 3 (4.00)                \\ 
& National Baseline \supercite{nationalbaseline} & 0.343 & 0.555 & 0.424 & 0.353 & 0.547 & 0.429 & / & / & / & / \\

\cline{2-11}
\\[-11pt]

& \trafficsafe 8B           & \multicolumn{1}{c}{0.622}              & \multicolumn{1}{c}{0.630}              & \multicolumn{1}{c|}{0.618}              & \multicolumn{1}{c}{0.640}              & \multicolumn{1}{c}{0.636}              & \multicolumn{1}{c|}{0.634}              & \multicolumn{1}{c}{0.756}              & \multicolumn{1}{c}{0.763}              & \multicolumn{1}{c|}{0.755}              & 2 (2.22)                \\
& \trafficsafe 70B          & \multicolumn{1}{c}{\textbf{0.630}}              & \multicolumn{1}{c}{\textbf{0.682}}              & \multicolumn{1}{c|}{\textbf{0.649}}              & \multicolumn{1}{c}{\textbf{0.648}}              & \multicolumn{1}{c}{\textbf{0.644}}              & \multicolumn{1}{c|}{\textbf{0.644}}              & \multicolumn{1}{c}{\textbf{0.760}}              & \multicolumn{1}{c}{\textbf{0.775}}              & \multicolumn{1}{c|}{\textbf{0.759}}              & \textbf{1 (1.00)}                \\ 
\midrule
\midrule
% \multicolumn{11}{c}{\textbf{Illinois Dataset}} \\
% \midrule
% \end{tabular}
% }

%     \end{subtable}
%     % \vspace{0.5cm} % Space between subtables

% \vspace{5mm}
%     \begin{subtable}{\textwidth}
%   \centering
%     \caption{Illinois dataset}
%   \label{tab:performance_IL}
%   \resizebox{1\textwidth}{!}{
%   \begin{tabular}{c|ccc|ccc|ccc|c}
%   \toprule[1.5pt]
% \multirow{3}{*}{\textbf{Model}} & \multicolumn{3}{c|}{\textbf{Injury}}              & \multicolumn{3}{c|}{\textbf{Severity}}             & \multicolumn{3}{c|}{\textbf{Type}}              & \multirow{3}{*}{\textbf{Rank}} \\ \cline{2-10}
%                        & \multicolumn{1}{c}{\textbf{Accuracy }} & \multicolumn{1}{c}{\textbf{Precision }} & \multicolumn{1}{c|}{\textbf{F1-score }} & \multicolumn{1}{c}{\textbf{Accuracy }} & \multicolumn{1}{c}{\textbf{Precision }} & \multicolumn{1}{c|}{\textbf{F1-score }} & \multicolumn{1}{c}{\textbf{Accuracy }} & \multicolumn{1}{c}{\textbf{Precision }} & \multicolumn{1}{c|}{\textbf{F1-score }} &                       \\ 

     \multirow{8}{*}{\textbf{Illinois}} &   RandomForest \supercite{randomforest}      & 0.462                                         & 0.554                                         & \multicolumn{1}{c|}{0.383}                                       & 0.430                                         & 0.452                                         & \multicolumn{1}{c|}{0.338}                                       & 0.610                                         & 0.670                                         & 0.632                                       &      3 (4.11)                 \\
                       &  AdaBoost \supercite{FREUND1997119}          & 0.403                                         & 0.183                                         & \multicolumn{1}{c|}{0.251}                                       & 0.318                                         & 0.147                                         & \multicolumn{1}{c|}{0.200}                                       & 0.109                                         & 0.083                                         & 0.083                                       &    8 (8.00)                   \\
                       &  CatBoost \supercite{prokhorenkova2019catboost}          & 0.457                                         & 0.543                                         & \multicolumn{1}{c|}{0.388}                                       & 0.454                                         & 0.446                                         & \multicolumn{1}{c|}{0.404}                                       & 0.535                                         & 0.656                                         & 0.579                                       &    4 (4.22)                   \\
                       &  DecisionTree \supercite{quinlan1986induction}      & 0.426                                         & 0.514                                         & \multicolumn{1}{c|}{0.410}                                       & 0.417                                         & 0.398                                         & \multicolumn{1}{c|}{0.361}                                       & 0.504                                         & 0.624                                         & 0.548                                       &   6 (5.33)                    \\
                       &  LogisticRegression \supercite{cox1958regression} & 0.413                                         & 0.439                                         & \multicolumn{1}{c|}{0.410}                                       & 0.360                                         & 0.385                                         & \multicolumn{1}{c|}{0.355}                                       & 0.379                                         & 0.477                                         & 0.400                                       &   7 (6.33)                    \\
                       &  XGBoost \supercite{10.1145/2939672.2939785}           & 0.442                                         & 0.575                                         & \multicolumn{1}{c|}{0.340}                                       & 0.405                                         & 0.419                                         & \multicolumn{1}{c|}{0.278}                                       & 0.678                                         & 0.694                                         & 0.683                                       &   5 (4.56)                    \\ 
                       & National Baseline \supercite{nationalbaseline} & 0.369 & 0.136 & 0.199 & 0.442 & 0.195 & 0.271 & / & / & / & / \\

                         \cline{2-11}
                         \\[-11pt]
                      &   \trafficsafe 8B       & 0.529                                         & 0.529                                         & \multicolumn{1}{c|}{0.533}                                       & \textbf{0.578}                                         & \textbf{0.584}                                         & \multicolumn{1}{c|}{\textbf{0.571}}                                     & 0.701                                         & \textbf{0.768}                                        & 0.721                                       &    2 (1.89)                   \\
                      &   \trafficsafe 70B      & \textbf{0.534}                                         & \textbf{0.587 }                                        & \multicolumn{1}{c|}{\textbf{0.543}}                                       & 0.554                                         & 0.561                                         & \multicolumn{1}{c|}{0.548}                                       & \textbf{0.727}                                         & 0.767                                        & \textbf{0.737}                                     &   \textbf{1 (1.44)}                        \\

    \midrule
    \midrule
\multirow{1}{*}{\textbf{North Carolina}} & \trafficsafe 8B (zero-shot) & 0.511 & 0.776 & \multicolumn{1}{c|}{0.468} & 0.549 & 0.638 & \multicolumn{1}{c|}{0.487} & 0.691 & 0.775 & 0.672 & / \\
\midrule
\midrule
\multirow{1}{*}{\textbf{Maine}} & \trafficsafe 8B (zero-shot) & 0.521 & 0.573 & \multicolumn{1}{c|}{0.457} & 0.542 & 0.582 & \multicolumn{1}{c|}{0.493} & 0.701 & 0.622 & 0.613 & / \\
   \bottomrule[1.5pt]
  \end{tabular}

  }

    \end{subtable}

    \label{tab:maintable}
\end{table}

%% file: discussion.tex
\section{Discussion}

\textbf{Deciphering traffic crash modeling as a linguistic learning task is a promising way for future safety research.} Most of the existing ML models for crash prediction typically treat various factors as independent numerical input variables \supercite{theofilatos2019comparing, bhuiyan2022crash}. However, this approach fails to capture information richness from the textual crash reports, such as detailed descriptions of behaviors, vehicle movements prior to the crash, and the traffic conditions. 
% Moreover, it further complicates the integration of multi-modal data as input. 
To address these issues, we employ an AI-expert cooperative prompt design approach to process diverse data types, including crash reports (textual), satellite and crash images (visual), and infrastructure characteristics (categorical), into a textual \dataset and use LLM for prediction. This transformation reframes the task of crash prediction into a text reasoning problem, enabling the use of LLMs to analyze and predict outcomes while preserving the rich, detailed textual information in crash reports, rather than reducing it to simplistic numerical representations. 
% By leveraging LLMs, the approach ensures that the rich, detailed textual information present in crash reports is preserved and integrated into the predictive process, rather than being reduced to simplistic numerical representations. 
As demonstrated by the results in Section \ref{sec:performance}, with our customization process, the \model outperforms all the baseline models, highlighting the advantages of reasoning through textual representations. Building on this, \attribution extends the framework by enabling conditional analysis of textual prompts, quantifying the contribution of specific factors to crash outcomes under various scenarios. As shown in Section \ref{sec:attribution}, this approach effectively identifies key contributors to crashes and high-risk scenarios, and offers data collection guidance for the iterative improvements in the future. 

% In general, the \trafficsafe framework provides a robust and interpretable solution for crash prediction and analysis, leveraging language-based reasoning to uncover actionable insights and guide evidence-based policymaking.

% Although the model already demonstrates strong performance, its capabilities can be further enhanced by incorporating additional data, such as sequential driver history for valuable context, and optimizing prompt design to explore diverse structures for improved performance. By embracing the richness of language and context, the \textit{TrafficSafe LLM} opens new avenues for understanding and predicting crash events, ultimately contributing to safer roads.

% trustworthiness, potential impact and application of the trustworthiness for decison making
\textbf{Providing reliable and interpretable predictions with quantifiable trustworthiness.} As illustrated in Figures \ref{fig:cs_wa} and \ref{fig:cs_il}, \model demonstrates a deep understanding of input–output correlations, yielding predictions whose accuracy increases alongside higher confidence scores. In all tasks across Washington and Illinois dataset, when the \model's confidence score exceeds 60\%, which captures over 70\% of the crash events under consideration. Furthermore, the confidence scores for fatal crash predictions are notably higher than those of other crash categories (60\% of model confident score leading to over 95\% of the real-world occurrence risk, see Figures \ref{fig:fatal_il_acc}, \ref{fig:fatal_il_conf}, and \ref{fig:fatal_il}). This trackable confidence-precision correlation can provide decision-makers with a robust tool for forecasting crashes under quantifiable uncertainty. Beyond predictive trustworthiness, the \attribution framework provides interpretable feature attribution by quantifying each feature’s contribution to the confidence score (see Section \ref{sec:method:attribution}). A higher feature contribution translates into a higher confidence score, which in turn yields greater prediction accuracy for the severe crash outcomes. Thus the factor with higher feature contribution value has higher impact to the model's prediction. For instance, alcohol-impaired driving (BAC > 0) increases the confidence score for severe crash predictions by more than 0.47, serving as a critical indicator for the likelihood of these severe crash outcomes.

\textbf{Identifying high-risk traffic crashes through conditional factors attribution even in unseen scenarios.} 
The \trafficsafe framework enables a detailed, sentence-level analysis of crash factors through conditional attribution, yielding critical insights into high-risk scenarios. In data-rich situations where sufficient data is available for each condition, \trafficsafe can rank the risk levels associated with various conditions, offering a prioritized list of scenarios that pose the highest danger. This capability supports targeted policy interventions by identifying specific conditions that substantially increase crash risks. For instance, as shown in Figure \ref{fig:5}a, driving in a work zone under sober conditions poses low level of risk; however, alcohol consumption in the same setting dramatically amplifies this risk, creating one of the most hazardous scenarios for severe crashes. This insight suggests potential policy interventions, such as mandatory BAC testing in work zones, to mitigate these risks. Likewise, Figure \ref{fig:5}c highlights that aggressive driving and impairment-related behaviors markedly increase the likelihood of serious or fatal outcomes, emphasizing the importance of driver education to discourage aggressive behavior and driving under the influence. 
\textbf{Moreover, \trafficsafe can be generalized to predict and understand data-sparse scenarios through "what-if" analysis, allowing hypothetical changes to specific conditions to be tested and their potential impact evaluated. }For example, while this study lacked sufficient data to comprehensively analyze the effects of user type (e.g., pedestrians or not pedestrians) or roadway type (e.g., freeway or not freeway), \trafficsafe provides a reliable mechanism to simulate and analyze such conditions. 

% By leveraging this approach, future research can explore the influence of these factors even in the absence of extensive datasets, offering valuable insights into scenarios that are underrepresented in the data.

% The \textit{TrafficSafe} framework facilitates a nuanced analysis of crash factors through conditional attribution, revealing critical insights into high-risk scenarios. By employing sentence-level feature contributions calculations, the framework identifies how specific contextual factors—such as driver behavior, BAC, roadway type—contribute to the crashes. For instance, as shown in Figure \ref{fig:5}c, aggressive driving and impairment-related behaviors consistently amplify the risk of serious and fatal outcomes. The framework’s ability to attribute risk under conditional scenarios enables policymakers to design evidence-based safety measures. For example, the analysis of serious and fatal crashes reveals that driving after drinking is the leading contributing factor. Moreover, driving within a work zone while under the influence of alcohol can significantly amplify the risk. By identifying such patterns, the model supports the development of scenario-specific safety guidelines, such as enhanced signage or stricter enforcement in work zones, which could mitigate risks effectively.

\textbf{Assessing the impact of data utility for improved future data collection and life-long learning.} Traffic crash data are inherently complex and multi-modal, making it crucial to identify which components are most informative and how critical they are for future traffic safety data collection. During the training stage, data attribution analysis revealed that unit information (e.g., driver behavior and vehicle details) and event information (e.g., vehicle movement and environmental conditions) exert the greatest influence on crash prediction performance. Specifically, for the \textit{Severity} task in the Illinois dataset, these features contributed 0.173 and 0.287, respectively, to the model’s prediction confidence (see Figure \ref{fig:5}e). These findings underscore the importance of prioritizing the collection of detailed, high-quality movement and behavior data in crash events, such as precise records of alcohol use, vehicle defects, vulnerable users' status, and road conditions. In contrast, although general information (feature contribution of 0.038) and infrastructure information (feature contribution of 0.019) remain valuable, their impact on some tasks is comparatively smaller. By directing data collection efforts toward gathering richer, more consistent information format in these critical safety domains, the accuracy of \trafficsafe can be further improved.

\textbf{Limitations of the proposed \trafficsafe.} A primary limitation relates to the handling of multi-modal data. In the \trafficsafe framework, satellite images were processed into textual descriptions and incorporated into prompts. While this approach offers flexibility, advancements in multi-modal foundation models and increasing research on integrating multi-modal data with LLMs present promising alternatives \supercite{zhang2024vision}. Leveraging specialized image encoders or utilizing multi-modal foundation models for processing image data are compelling directions. Another potential limitation lies in the efficiency of model training and attribution. Fine-tuning LLMs and computing feature contributions have always required substantial resources and time. Although we employed LoRA fine-tuning and a stratified sampling technique to enhance efficiency \supercite{10.1145/3588728}, implementing the complete framework still demands significant resources. This poses certain limitations when resources are scarce or in situations demanding rapid model deployment.  

%% file: method.tex
\section{Methods}

\subsection{\textit{TrafficSafe Event} Dataset Construction}

As introduced in Section \ref{sec:raw_data}, the raw crash data is multi-modal, and integrated from various sources. To adapt the raw data for LLMs' fine-tuning process, we employ the feature engineering and textualization process to generate textual inputs. In this section, we will discuss the formats of raw data and the textualization process (see Extended Data Figure \ref{fig:data_processing}).

\subsubsection{Raw Data}
\label{sec:methods:raw_data}
The raw crash data used in this study was obtained from the HSIS \supercite{hsis} and Google Maps \supercite{googlemaps}. Data from the HSIS, sourced from multiple systems, encompasses a variety of formats, including categorical, numerical, and textual. In total, four main datasets were used:

\begin{itemize}
    \item \textbf{Crash data.} This dataset captures the essential spatio-temporal and contextual attributes of each crash. It includes crash date, time, day of the week, and month, along with location details such as route number, milepost, and the surrounding area’s classification (e.g., rural or urban). Higher-level planning attributes (e.g., roadway and functional classifications, intersection-related indicators) are also recorded. In addition, it documents the dynamic circumstances leading up to the event, including the number of vehicles and pedestrians involved, vehicle travel directions (increasing or decreasing milepost), and any maneuvers performed (e.g., lane changes, straight-line movement).
    \item \textbf{Infrastructure data.} This dataset details the physical and infrastructural features of the crash site. Key elements include the type of road surface (e.g., asphalt or concrete), average annual daily traffic (AADT), posted speed limits, and access control mechanisms. It also encompasses dimensions such as total road width, right and left shoulder widths, and median width (including median barriers if present), as well as road surface conditions (e.g., dry or wet) and ambient lighting at the time of the crash (e.g., daylight or dusk).
    \item \textbf{Vehicle data.} This dataset consolidates information on the vehicles involved in each crash, including vehicle type (e.g., passenger car or truck), intended use (e.g., commercial or private), mechanical condition (e.g., defects), and relevant driver actions (e.g., lane changes or stopping). Additional information on airbag deployment and occupant ejection status provides further granularity.
    \item \textbf{Person data.} This dataset compiles information about individuals involved in the crash, detailing demographic characteristics such as age, gender, and seating position. It also includes the use of safety equipment (e.g., seat belts or helmets) and any contributing factors, such as driver distraction or impairment.
\end{itemize}

The satellite images obtained from Google Maps serve as a supplementary data source to complement the HSIS dataset. Details of the integration process are provided in Section \ref{sec:methods:Feature_engineering}. Overall, we collect 16,188 crash events data from Washington State and 42,715 events from Illinois State for further analysis.

\subsubsection{Feature Engineering and Textulization of Crash Data}
\label{sec:methods:Feature_engineering}

To adapt the multi-modal data to the input of LLMs, we followed the following process to generate textual prompt from raw data entry:

\begin{itemize}
    \item \textbf{Data mapping and organization.} For each crash, we associated the crash report with the involved vehicles and individuals using the crash ID, thus obtaining descriptions of the crash and the persons involved. The route ID and milepost were used to identify the specific road segment where the crash occurred, allowing us to gather related road and environment information from infrastructure data. The integrated data was then systematically organized into four categories: general information, infrastructure information, event information, and unit information, aligning with the components outlined in Section \ref{sec:results:Constructing}.
    \item \textbf{Satellite images textualization.} The HSIS datasets provide GPS coordinates for crash locations in Washington and Illinois. To address missing information such as the number of road lanes, high-resolution satellite images (512 × 512 pixels at a zoom level of 19) were retrieved using these GPS coordinates via the Google Maps API. These images supplement the crash dataset with crucial infrastructure and environmental context. Descriptive textual annotations were generated from the satellite images using GPT-4, filling key gaps in the original dataset. These annotations include information such as the number of lanes at the crash site, whether the crash occurred at an intersection, and whether the surrounding area is residential.
    % To supplement the infrastructure and environmental information, we obtained satellite images based on GPS coordinates and used GPT-4o to generate images' description, such as "the crash is located at an intersection in a residential area". 
    \item \textbf{Dimensionality reduction.} Raw data include abundant attributes with rich and varied descriptions. However, some features suffer from insufficient distinction between attribute values due to the original classification's complexity. To address this, we performed dimensionality reduction on these attributes by combining domain experts' insights with GPT-4o clustering results. For example, similar classifications like "pedalcyclist struck by vehicle" and "pedalcyclist strikes vehicle" were clustered under a broader category such as "pedalcyclist collisions". This process generalized the data and reduced redundancy.
    \item \textbf{Prompt generation using AI-expert textualization method.} To generate logically coherent and continuous textual data suitable for LLM training, we transformed each category of data into text format using GPT-4. All data are organized as key-value pairs and we get four parts of the key-value pairs for each event case. Then GPT-4o is used to generate the text prompt for each section of the key-value pairs individually. For each part, we apply straightforward prompt to GPT-4o, such as "\textit{Please translate a python dictionary to paragraph, act as a crash data interpreter.}". The text content is extracted from GPT-4o's response for each part consisting of approximately 100 words. By linking four parts of text, we obtain a comprehensive textual description for each crash event case. The detailed process is shown in Extened Data Figure \ref{fig:ai_expert}.
    % These textual outputs were then structured as input and output for the LLMs. 
    %Detailed information regarding the input and output design for an individual crash event is provided in Section \ref{sec:input_output}.
    % \item \textbf{Expert review.} To ensure the accuracy of the generated text contents, during the labeling process, we randomly select 5\% of the data across different months, locations, genders, and infrastructure conditions for review by human experts. This helps verify that the generated content is reasonable and aligns with real-world scenarios.
    
\end{itemize}

\subsubsection{Define Prediction Targets}
\label{sec:methods:targets}

We select three variables as the prediction targets: \textit{Injury}, \textit{Severity}, and crash \textit{Type}. The three targets are defined as:

\begin{itemize}
    \item The \textit{Injury} $n_i^\mathcal{D} \in \{f(l)|l=0,1,2,\cdots\}$, where i denotes the $i$-th data in the dataset, $\mathcal{D} \in \{\mathcal{W}, \mathcal{I}\}$ denotes the Washington dataset $\mathcal{W}$ or the Illinois dataset $\mathcal{I}$, $l$ represents the number of people injured,  and $f(l)$ denotes the label when the injured people is $l$.
    \item The \textit{Severity} $s_i^\mathcal{D} \in \{ S_k|k=1,2,\cdots\ \}$ (define on the KABCO scale\footnote{\url{https://highways.dot.gov/media/20141}}), where $S_k$ is the $k$-th level of crash severity.
    \item The \textit{Type} $ t_i^\mathcal{D} \in \{ T^\mathcal{D}_k|k=1,2,,\cdots\ \}$, where $T^\mathcal{D}_k$ is the $k$-th label of crash type in dataset $\mathcal{D}$.
\end{itemize}

We utilize these three variables to describe the crash result \(\text{CR}_i^{\mathcal{D}}\). The crash outcome can be presented in the following format: $\text{CR}_i^{\mathcal{D}} = (n_i^\mathcal{D}, s_i^\mathcal{D},t_i^\mathcal{D})$. For numerical variables, the function \( f(l) \)  describes the number of people injured in crash as follows: "zero" if \(l=0\), "one" if \(l=1\), "two" if \(l=2\), and "three and more than three" if \(l \geq 3\), the values for $S_k$ and $T^\mathcal{D}_k$ are provided in the Supplementary Table 2 and Supplementary Table 3.

\subsection{\model}

\label{sec:methods:model}

We fine-tune \model by adapting LLaMa 3.1~\supercite{meta2024llama} to crash prediction tasks to enhance the LLMs' capabilities in interpreting crash data, identifying critical factors, and conducting feature attribution analysis to offer insights for crash prevention. In this section, we will introduce detailed information of the fine-tuning process. 

\subsubsection{Construct Training Data for LLMs}
In the training of LLMs, a single input consists of three components: the system prompt, the user prompt, and the target prompt. The system prompt introduces the task, for example: "\textit{You are a helpful assistant designed to predict the severity of a traffic crash ...}”. The user prompt comprises the four content parts detailed in Section \ref{sec:methods:Feature_engineering} for each case. The target prompt represents the expected output. Examples of these prompts are shown in Extended Data Figure \ref{fig:WA_prompt}, Extended Data Figure \ref{fig:IL_prompt}, and Supplementary Section 3. We tokenize the text inputs using LLaMA 3.1's tokenizer.

\subsubsection{Additional Special Tokens for Classification}
To adapt the LLM as a crash classifier, additional tokens have been incorporated into the tokenizer's vocabulary, and the detailed crash attributes categories are listed in Supplementary Table 2 and Supplementary Table 3. Specifically, for predicting the number of people \textit{Injuries} of Washington dataset and Illinois dataset, we have introduced four special tokens: \texttt{<ZERO>}, \texttt{<ONE>}, \texttt{<TWO>}, and \texttt{<THREE AND MORE THAN THREE>}. Similarly, for predicting the Crash \textit{Severity} of Washington dataset and Illinois dataset, we use five additional tokens: $S_k$, where $1 \leq k \leq 5$, corresponding to different levels of severity. The \textit{Type} task differs slightly between the Washington and Illinois datasets. For Washington datasets, we utilize 14 special tokens: $T_k^{\mathcal{W}}$, where $1\leq k \leq 14$, each representing a specific crash type. For Illinois datasets, we utilize 16 special tokens: $T_k^{\mathcal{I}}$, where $1\leq k \leq 16$. The parameters of the input and output embedding layers are set as trainable, enabling the model to align the representations of these special tokens with the existing embedding space.

\subsubsection{Supervised Fine-tuning}
\label{sec:supervised_finetuning}
During the fine-tuning phase, the traffic forecasting task is framed as a next-token generation task. This process can be described as:
\begin{equation}
   p_{\theta}(T_i)=\prod_{j=1}^{|T_i|}{p_{\theta}(t_j^{(i)} \vert t_{1}^{(i)}, \cdots , t_{j-1}^{(i)}}),\vspace{-1mm}
\label{eq: autoregressive}
\end{equation}
where $T_i$ is the $i$-th item in the training data, $p_{\theta}$ is the LLM, $t_j^{(i)}$ denotes the $j$-th token in $T_i$. By maximizing the likelihood $ p_{\theta}(T)=\prod_{i=1}^{N}{p_{\theta}(T_i)}$, the LLM's parameters are learned. Both the system prompt and the user prompt are masked for loss computation during training. We also used uniform data sampling strategy during the training process to facilitate the convergence of \model \supercite{du2024advancing}. Through this process, the model learns to make prediction for a traffic crash. 

\subsubsection{Data Split}
\label{sec:data_split}
We split the Washington and Illinois dataset into training, validation, and test set in a 7:1.5:1.5 ratio. Since the Washington dataset contains relatively few crash events per year, we utilized as many reports as possible to ensure sufficient training data. However, the data distribution across different classes is highly imbalanced. For example, in the crash severity prediction task in Washington dataset, the ratio of $\#S_1/\#S_5$ is nearly 100:1, where $\# S_k$ is the number of data with label $S_k$. The imbalanced data distribution presents a significant challenge for the model's training and evaluation. In Section \ref{sec:supervised_finetuning}, we used uniform sampling strategy to train model on this unbalanced data. Similarly, to facilitate the model's evaluation, for the validation set and test set, we removed most of the data with crash severity category of $S_1$. Specifically, after processing, the dataset consisted of 16,188 records, with 11,332 used for training, 2,428 for validation, and 2,428 for testing. To balance the validation and test set for better evaluation, we removed 1428 $S_1$ data and used 1000 remaining data for validation set and test set separately. Compared with the Washington state, more crash records can be used in Illinois state to generate dataset. As a result, we were able to balance all subsets, including the training, validation, and test sets. Ultimately, the Illinois dataset comprised 42,715 records, with 29,307 used for training, 6,704 for validation, and 6,704 for testing.

\subsubsection{Evaluation Metrics}
\label{sec:methods:metrics}
In evaluating the model performance as a classification task, we employ weighted accuracy, precision, and F1-score as metrics. In the context of a classification task, we have four notations, True Positive ($TP$), True Negative ($TN$), False Positive ($FP)$, False Negative ($FN$). Using these notations, we can represent the metrics as follows:
\begin{itemize}
    \item Accuracy is one of the most commonly used measures for the classification performance, and it is defined as a ratio between the correctly classified samples to the total number of samples as follows:
    \vspace{0.2em}
    \begin{equation}
    \text{Accuracy}=\frac{TP+TN}{TP+TN+FP+FN}
    \end{equation}
    \item Precision represents the proportion of positive samples that were correctly classified to the total number of positive predicted samples, which reflect the performance of the prediction:
    \vspace{0.2em}
    \begin{equation}
    \text{Precision}=\frac{TP}{TP+FP}
    \end{equation}
    % \item $Recall$ is used to measure the fraction of positive patterns that are correctly classified. It can be calculated using formula:
    % \vspace{0.2em}
    % \begin{equation}
    % Recall=\frac{TP}{TP+FN}
    % \end{equation}
    \item F1-score  combines results on precision and recall. It is the harmonic mean of precision and recall, which can be calculated using formula:
    \vspace{0.2em}
    \begin{equation}
    \text{F1-score}=\frac{2}{\text{Precision}^{-1}+\text{Recall}^{-1}}=2\cdot(\frac{\text{Precision} \cdot \text{Recall}}{\text{Precision}+\text{Recall}})
    \end{equation}
    where $\text{Recall}=TP/(TP+FN)$.
\end{itemize}

\subsubsection{Adopted Baselines}
\label{sec:methods:baselines}
We follow the recent literature~\supercite{ahmed2023study} and also adopt XGBoost~\supercite{10.1145/2939672.2939785}, Random forest (RF)~\supercite{randomforest}, Decision Trees (DT)~\supercite{quinlan1986induction}, Adaptive boosting (AdaBoost)~\supercite{freund1999short}, LogisticRegression (LR)~\supercite{cox1958regression}, Categorical boosting (CatBoost)~\supercite{prokhorenkova2018catboost}, and National Average~\supercite{nationalbaseline} as compared baselines. Building upon these foundational models, we particularly focus on enhancing their predictive capabilities through advanced techniques and parameter optimization. The detailed descriptions of these models are listed as follows:
\begin{itemize}
    \item \textbf{XGBoost} is a scalable and distributed gradient-boosting framework that constructs an ensemble of decision trees by minimizing a regularized loss function. It uses second-order gradients for optimization and includes features like shrinkage, column subsampling, and tree pruning to improve accuracy and prevent overfitting~\supercite{10.1145/2939672.2939785}.

    \item \textbf{AdaBoost}~\supercite{FREUND1997119} is an iterative boosting method that sequentially trains weak classifiers (e.g., decision stumps) and assigns higher weights to misclassified instances in subsequent iterations. The final prediction is determined by a weighted majority vote of all classifiers.

    \item \textbf{Random Forest (RF)}~\supercite{randomforest} builds an ensemble of decision trees by randomly sampling both features and data points (via bootstrap aggregation). The aggregated (voted) output of these diverse trees reduces variance and provides robust performance across a variety of tasks.
    % In this study, Random Forest serves as a baseline model, with its hyperparameters optimized using Bayesian optimization techniques.
    \item \textbf{Decision Trees (DT)}~\supercite{quinlan1986induction} recursively split the feature space based on selected thresholds, forming a hierarchical tree structure that is easy to interpret. Although they can capture complex interactions, DTs are prone to overfitting if not properly regularized.
    \item \textbf{Logistic Regression (LR)}~\supercite{cox1958regression} models the probability of a binary outcome through a linear combination of input features passed through the logistic function. Coefficients are typically estimated via maximum likelihood, providing a simple yet effective approach for classification.
    \item \textbf{CatBoost}~\supercite{prokhorenkova2018catboost} is a gradient-boosting algorithm that efficiently handles categorical features through techniques such as ordered boosting and gradient-based one-hot encoding. By systematically reducing target leakage in encoding, it achieves high predictive accuracy while mitigating overfitting in heterogeneous datasets.
    \item \textbf{National Average}~\supercite{nationalbaseline} predicts crash severity distributions using calibrated Severity Distribution Functions (SDFs). It incorporates road design, traffic control, and crash data to estimate probabilities for different severity levels via a multinomial logit model.
\end{itemize}

For these models, the Bayesian optimization method (\textit{BayesSearchCV}) is used to facilitate the identification of optimal hyperparameters, such as \textit{max\_depth} and \textit{learning\_rate}. The details of the hyperparameters settings of these models are shown in Supplementary Section 4.

\subsection{\attribution}
\label{sec:method:attribution}
To identify the feature contribution of each factor to the prediction results, this paper introduces and adapts the concept of Shapley values \supercite{bordt2023shapley}. In this section, we first explain the calculation process of Shapley values and subsequently propose a novel sentence-level feature contibutions calculation method based on Shapley theory for attributing factors in LLMs.

\subsubsection{Definition of Shapley Value}\label{sec:shapley_def}
\label{sec:shapley}
Shapley value is a concept from cooperative game theory that has been widely adopted in machine learning to interpret model predictions \supercite{chen2022explaining}. It provides a way to fairly allocate the contribution of each feature to the outcome of a predictive model. In essence, the Shapley value quantifies how much each feature contributes to a prediction by considering all possible combinations of features. Formally, the Shapley value $\varphi$ of a feature (or player) $i$ in a cooperative game is defined as:
\begin{equation}
\label{eq:shapley}
    {\varphi}_{i} = \sum_{S \subseteq N \setminus \{i\}} \frac{|S|!(n - |S| - 1)!}{n!} \Bigl[v(S \cup \{i\}) - v(S)\Bigr],
\end{equation}
where $N = \{1, 2, \dots, n\}$ is the index set of $n$ features, $S$ is a subset of $N$, and $v(S)$ is the utility of the subset $S$, which represents a measurable value, such as accuracy or prediction score, achieved by the model using only the subset $S$ of features.

The Shapley value is utilized in both the training and inference stages in \trafficsafe. During the training stage, it quantifies the contributions of four primary categories of information: general information, infrastructure information, event information, and unit information. During the inference stage, the Shapley value is applied to assess the contributions of individual sentences to the prediction outcomes. The specific methodologies and implementation details are outlined in the subsequent sections.
% and function $v(S)$ indicates the payoff of a coalition $S$ (which can be interpreted as a metric such as the softmax probability or the model's accuracy). Summed all $SV_i$ together, these contributions produce a fair attribution of the total payoff to each feature, ensuring that every feature is credited precisely for its expected impact on the final outcome.
% In our research, we computed two types of Shapley values: Data Shapley in training stage and Data Shapley in inference stage. In the following sections, we will provide a detailed explanation of their implementation and results.

\subsubsection{Feature Contributions at the Training Stage}
\label{sec:training-shapley}
% Based on the categorization in Section \ref{sec:results:Constructing}, the prompts are split into four parts to evaluate how different components in the training set influence the model during training, which are  general information, infrastructure information, event information, and unit information.  According to Equation \eqref{eq:shapley}, Shapley value of each component during the training stage $SV_{i}^t$ is calculated, 

The Shapley value is utilized to assess the influence of different components in the training set on the model during training. As outlined in Section~\ref{sec:results:Constructing}, the $j$-th prompt $p_j$ in the dataset $P$ is divided into five parts: $c_0$: system prompt (i.e. \textit{"You are a helpful assistant designed to predict the severity of a traffic crash ..."}), $c_1$: general information, $c_2$: infrastructure information, $c_3$: event information, and $c_4$: unit information. We denote $p_j(k)$ as the $c_k$ portion of $p_j$. Given an index set $S$, we can construct a variant $p_j(S)$ by concatenating the parts in $S$. For example, if $S = \{0, 1, 2\}$, then $p_j(S)$ contains $c_0$, $c_1$, and $c_2$. Formally,
\begin{equation}
\label{eq:train_shapley}
p_j(S) = \text{concat}_{k \in S}\, p_j(k),
\end{equation}
where \(\text{concat}\) denotes concatenation. The resulting dataset based on $S$ is $P(S) = \{ p_j(S) \mid j=0,1,\dots,L \}$, where $L$ is the dataset size.

Referring to Equation~\eqref{eq:shapley}, the contribution of part $c_i$ at training, ${\varphi}_{i}^{\text{train}}$, is
\begin{equation}
   {\varphi}_{i}^{\text{train}}=\sum_{S \subseteq N \setminus \{i\}} \frac{|S|!(n - |S| - 1)!}{n!} \cdot \Bigl[v\bigl(P(S \cup \{0, i\})\bigr) - v\bigl(P(S \cup \{0\})\bigr)\Bigr],
\end{equation}
where $N=\{1,2,3,4\}$ indexes the four content parts, and $v(P(S))$ is a performance metric (e.g., accuracy, F1-score) obtained after retraining the model only on prompts in $P(S)$.

\subsubsection{Sentence-level Feature Contributions at the Inference Stage}
\label{sec:sentence-shapley}

Unlike traditional machine learning models that primarily handle fixed-length feature vectors, LLMs process variable-length text sequences as input\supercite{chen2023algorithms}. This characteristic makes commonly used Shapley value approximation methods, such as KernelSHAP \supercite{NIPS2017_8a20a862} and DeepSHAP, less applicable to LLMs. Recent approaches like TokenSHAP \supercite{goldshmidt2024tokenshap} and TransSHAP \supercite{kokalj-etal-2021-bert} have been proposed to address this by decomposing input text into tokens and computing Shapley values at the token level. However, applying token-level Shapley value computation to \model introduces two primary challenges: 1) Computational limitations. The computational complexity of Shapley values is exponential in the number of players. In our \model, with an input size of approximately 500 tokens, large-scale computation of token-level Shapley values for crash data becomes impractical. 2) Limited interpretability. Decomposing the prompt at the token level disregards inter-token dependencies, and the arbitrary masking or replacement of tokens can lead to semantic ambiguity and contextual shifts. These issues hinder a precise understanding of how individual features contribute to predictions. Moreover, paragraph-level analysis is too coarse for detailed attribution, since it can merge distinct features into a single category  (e.g., driver and vehicle details under “unit information”).

% In the following parts will detail the implementation of this framework.By segmenting the prompt into semantically coherent sequences and calculating Shapley values for each sequence, this approach not only reduces the number of features required for computation but also enhances the interpretability by grounding Shapley values in clearly defined semantic units.

To overcome these limitations, we propose a sentence-level feature contributions calculation method for inputs of LLMs, which proceeds as follows:

\begin{itemize}
    \item \textbf{Sentence segmentation.} The prompts are segmented using delimiters (e.g., commas "," or periods ".") to produce sentence-level units.
    \item \textbf{Feature groups annotation.} GPT-4o is used to group and label these sentences (see Figure \ref{fig:Severity_single_shap} for the groups' content). Each group is represented as $c_k$, where $k \in N' = \{1,2,3,\dots\,n \}$. For the Washington dataset, $n=14$, while for the Illinois dataset $n=12$. Given index set \(S' \subseteq N' \setminus \{\, i \}\), we can construct the the prompt $p_j(S')$ similar to the process Equation \eqref{eq:train_shapley}. The dataset built upon $S'$ can be written as $P(S') = \{ p_j(S')|j=0,1,2,\cdots,L \}$, where $L$ is the length of the dataset $P$.
    \item \textbf{Feature contributions calculation based on the feature groups.} Based on the constructed dataset, the feature contribution for the $i$-th sentence-group ${\varphi}_{i}^{inf}$ can be calculated as:

    \begin{equation}
        \label{eq:inference_shapley}
        {\varphi}_{i}^{inf}=\sum_{S' \subseteq N' \setminus \{i\}} \frac{|S'|!(n - |S'| - 1)!}{n!} \cdot \left[ p_{\theta}(P(S' \cup \{0,i\})) - p_{\theta}(P(S' \cup \{0\})) \right]
    \end{equation}
where $p_{\theta}$ is the LLM that returns the predicted probability of the target. A higher $\varphi_i^{\text{inf}}$ indicates a greater contribution of the $i$-th sentence group to the model’s confidence. To reduce computational overhead, we adopt a stratified sampling--based Shapley estimation method using complementary contributions~\supercite{10.1145/3588728}.

% where, \( p_{\theta} \) represents the LLM, which returns the predicted probability of the targets. Using this calculation, \( {\varphi}_{i}^{\text{inf}} \) quantifies the contribution of the \( i \)-th sentence group to the prediction's confidence. Specifically, a higher \( {\varphi}_{i}^{\text{inf}} \) indicates a greater contribution of the \( i \)-th sentence group to the confidence score of the prediction. To enhance algorithmic efficiency, we applied a Shapley value estimation approach utilizing a stratified sampling technique based on complementary contributions, as outlined in Zhang's work \supercite{10.1145/3588728}. 
\end{itemize}

%% file: extended_data.tex
\section{Extended Data}

% \captionsetup{labelformat=empty, labelsep=none}
% \captionsetup[figure]{labelformat=nolabelcolon, labelsep=none}
% \captionsetup[sub]{labelformat=nolabelcolon, labelsep=none}
\captionsetup[figure]{labelformat=default, labelsep=colon, name=Extended Data Figure}% change to Supplementary Figure
\captionsetup[table]{labelformat=default, labelsep=colon, name=Extended Data Table}
\setcounter{figure}{0}

\begin{figure}[!h]
    \centering
    \includegraphics[width=0.95\linewidth]{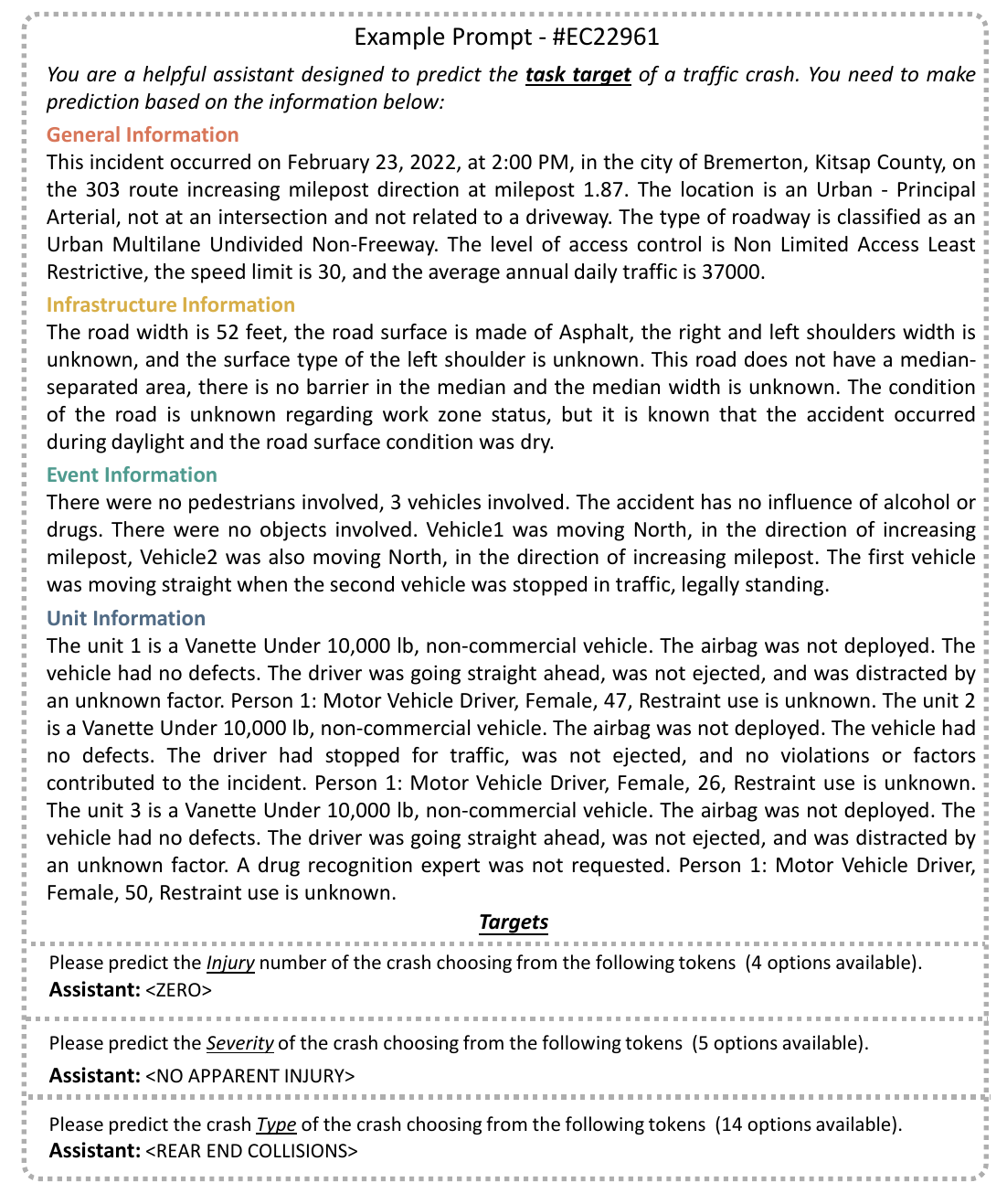}
    \vspace{-0.5cm}
    \caption{\textbf{A Crash Event Prompt Example from Washington Dataset.} }
    \label{fig:WA_prompt}
\end{figure}

\newpage

\begin{figure}[!h]
    \centering
    \includegraphics[width=0.95\linewidth]{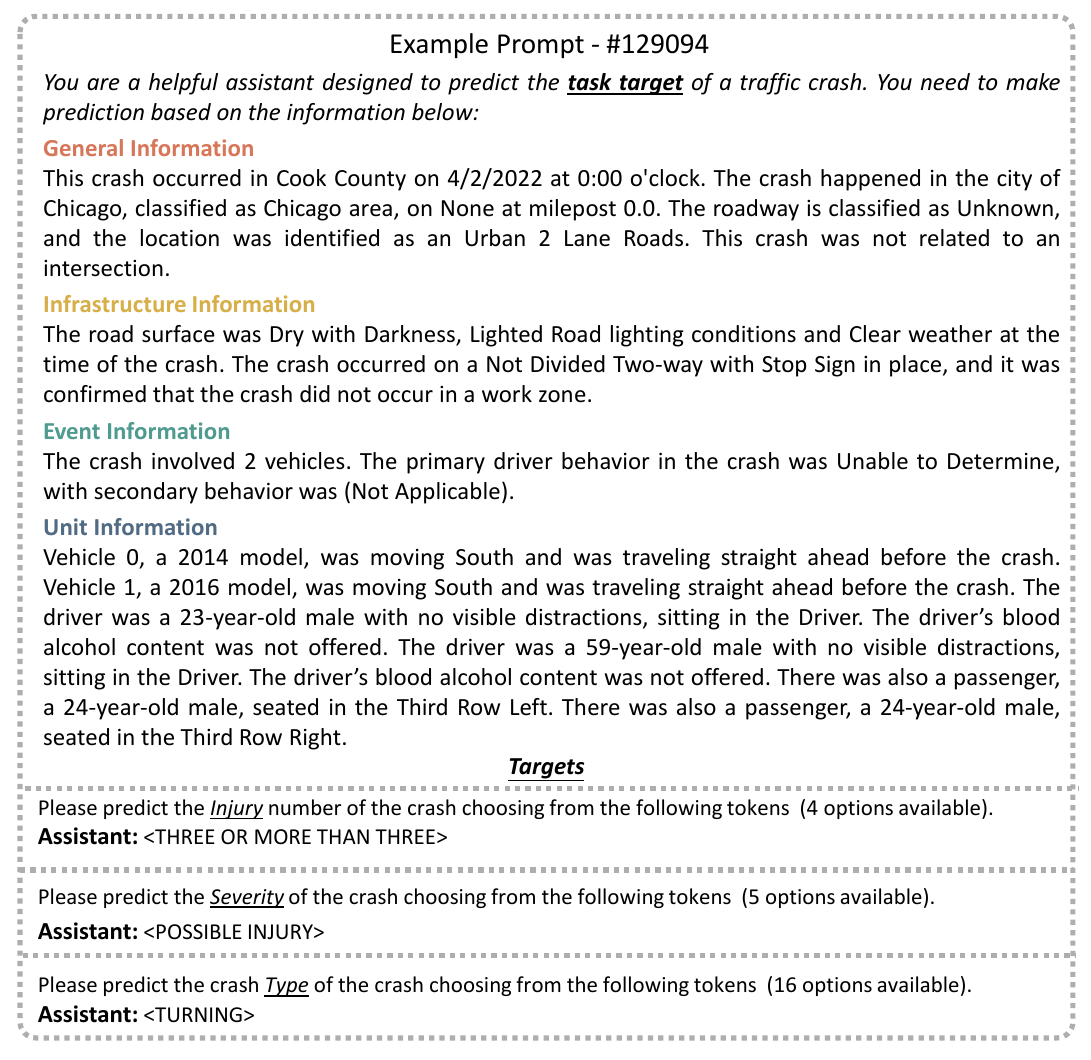}
    \vspace{-0.5cm}
    \caption{\textbf{A Crash Event Prompt Prompt Example from Illinois Dataset.}}
    \label{fig:IL_prompt}
\end{figure}

\newpage
\begin{figure}[!ph]
    \centering
    % \stepcounter{extfig}
    \begin{minipage}{.48\textwidth}
        \centering
        \includegraphics[width=\linewidth, height=.7\textheight]{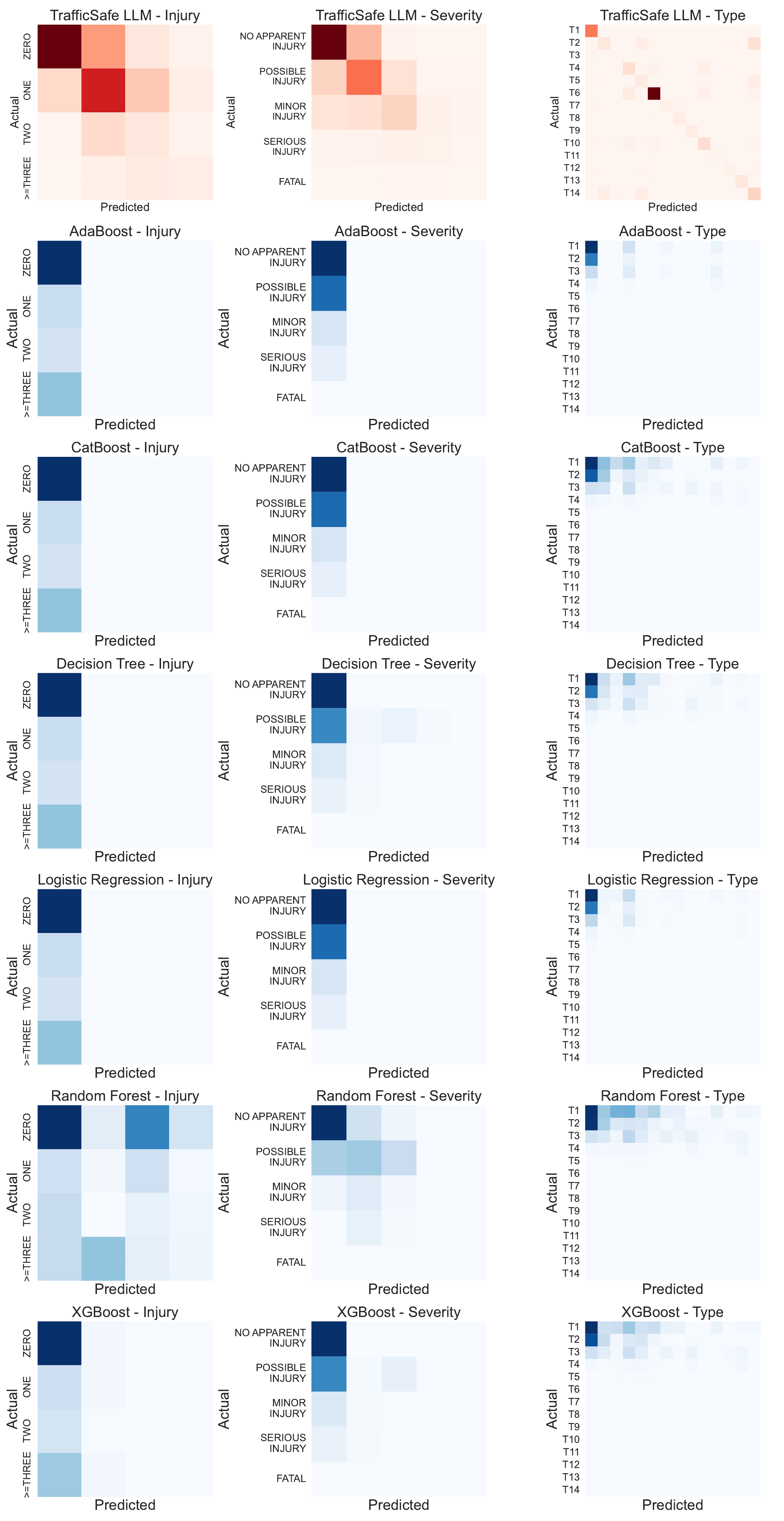}
        \subcaption{}
        \label{fig:extended_cm_WA}
    \end{minipage}%
    \hspace{0.02\textwidth}
    \begin{minipage}{.48\textwidth}
        \centering
        \includegraphics[width=\linewidth, height=.7\textheight]{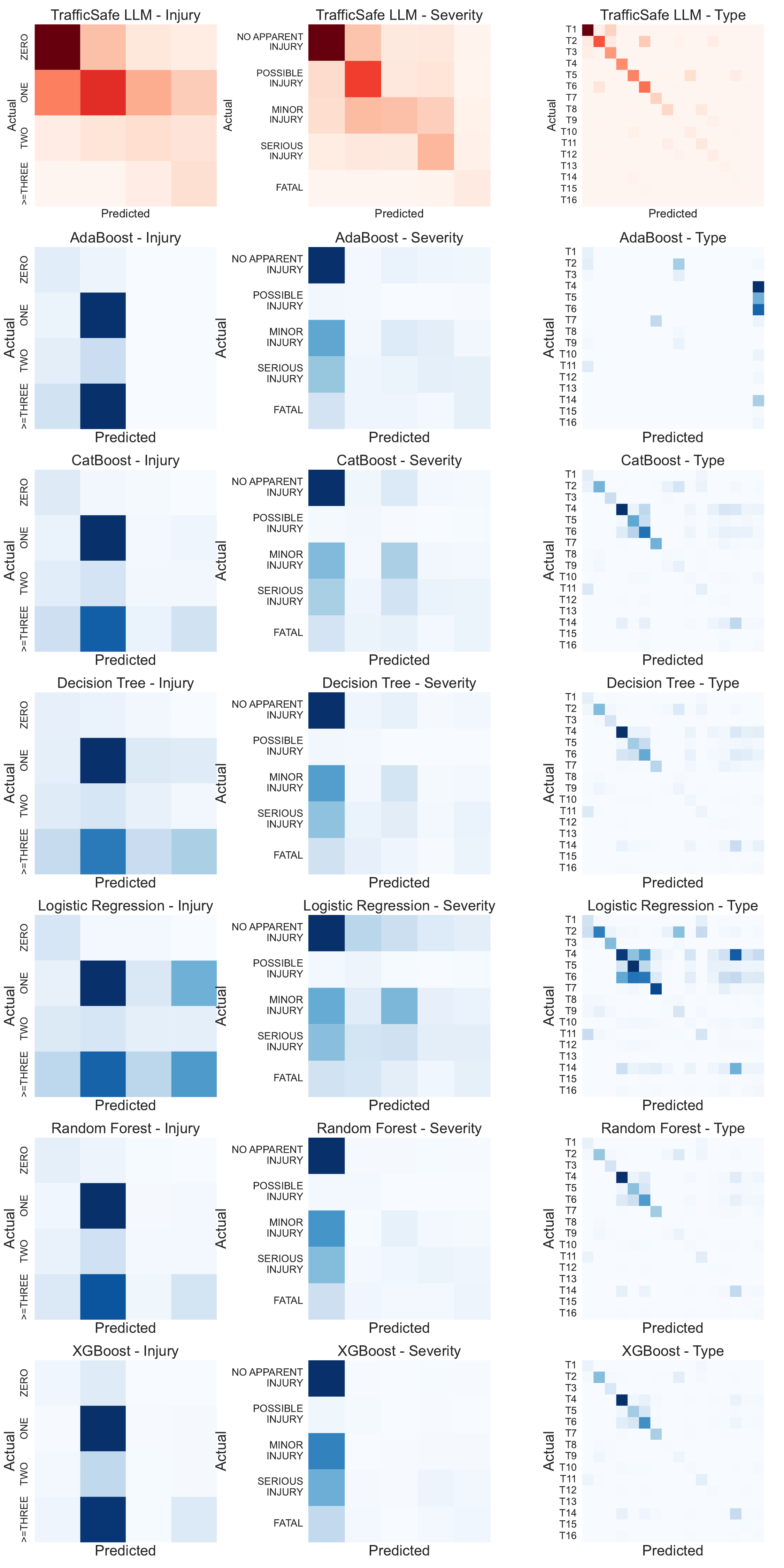}
        \subcaption{}
        \label{fig:extended_cm_IL}
    \end{minipage}
    \caption{\textbf{The Confusion Matrix for \model and the Traditional Methods in (a) Washington Dataset and (b) Illinois Dataset.} }
    \label{fig:extended_cm}
\end{figure}

\newpage
\begin{figure}[!ph]
    \centering
    \includegraphics[width=1\linewidth]{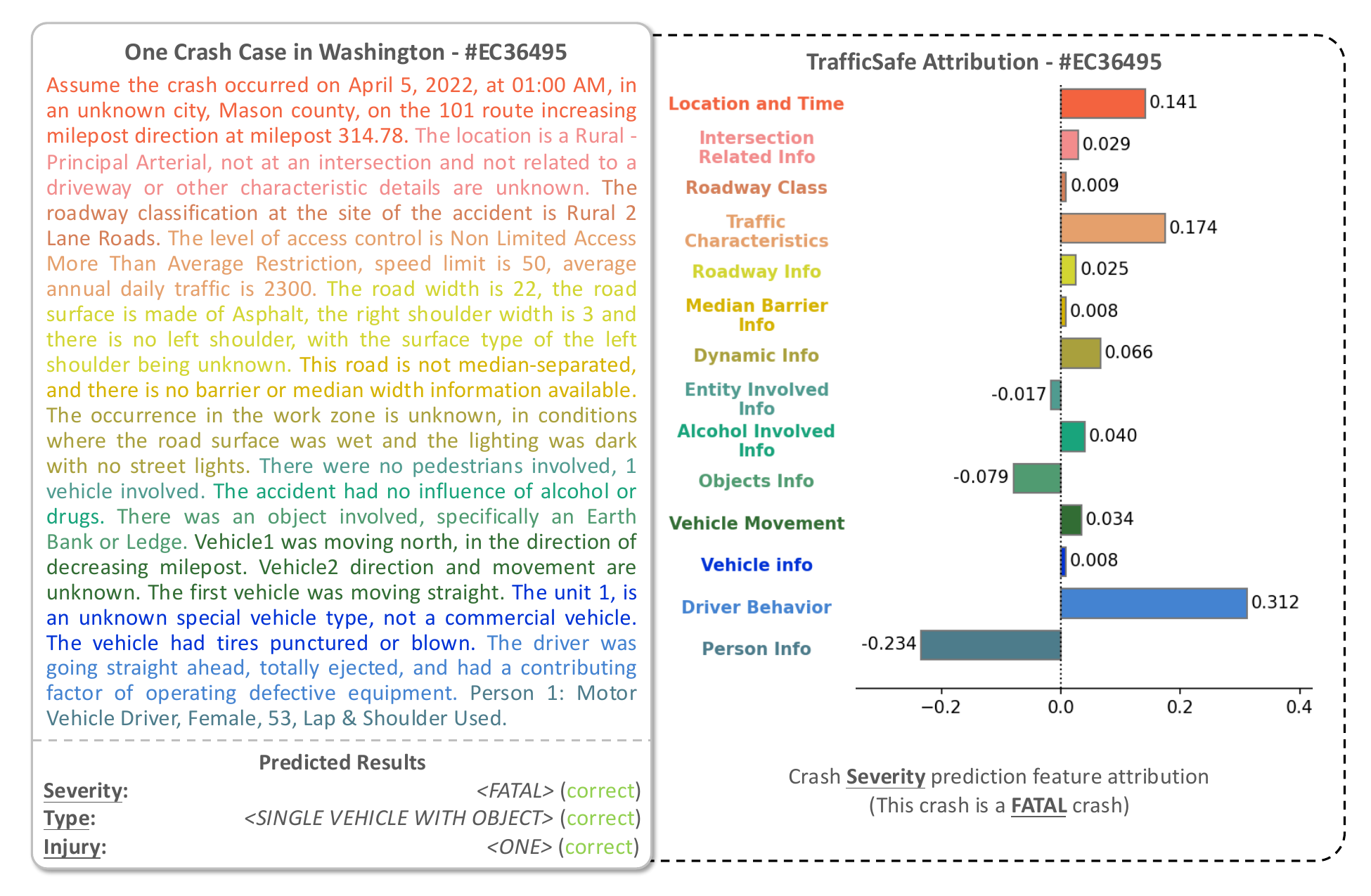}
    \vspace{-0.2cm}
    \caption{
        \textbf{
    One Example of Sentence-based Feature Attribution Results for A Crash Resulting in \textit{Fatal} in Washington Dataset.}
    }
    \label{fig:WA_FA2}
\end{figure}

\begin{figure}[!ph]
    \centering
    \includegraphics[width=1\linewidth]{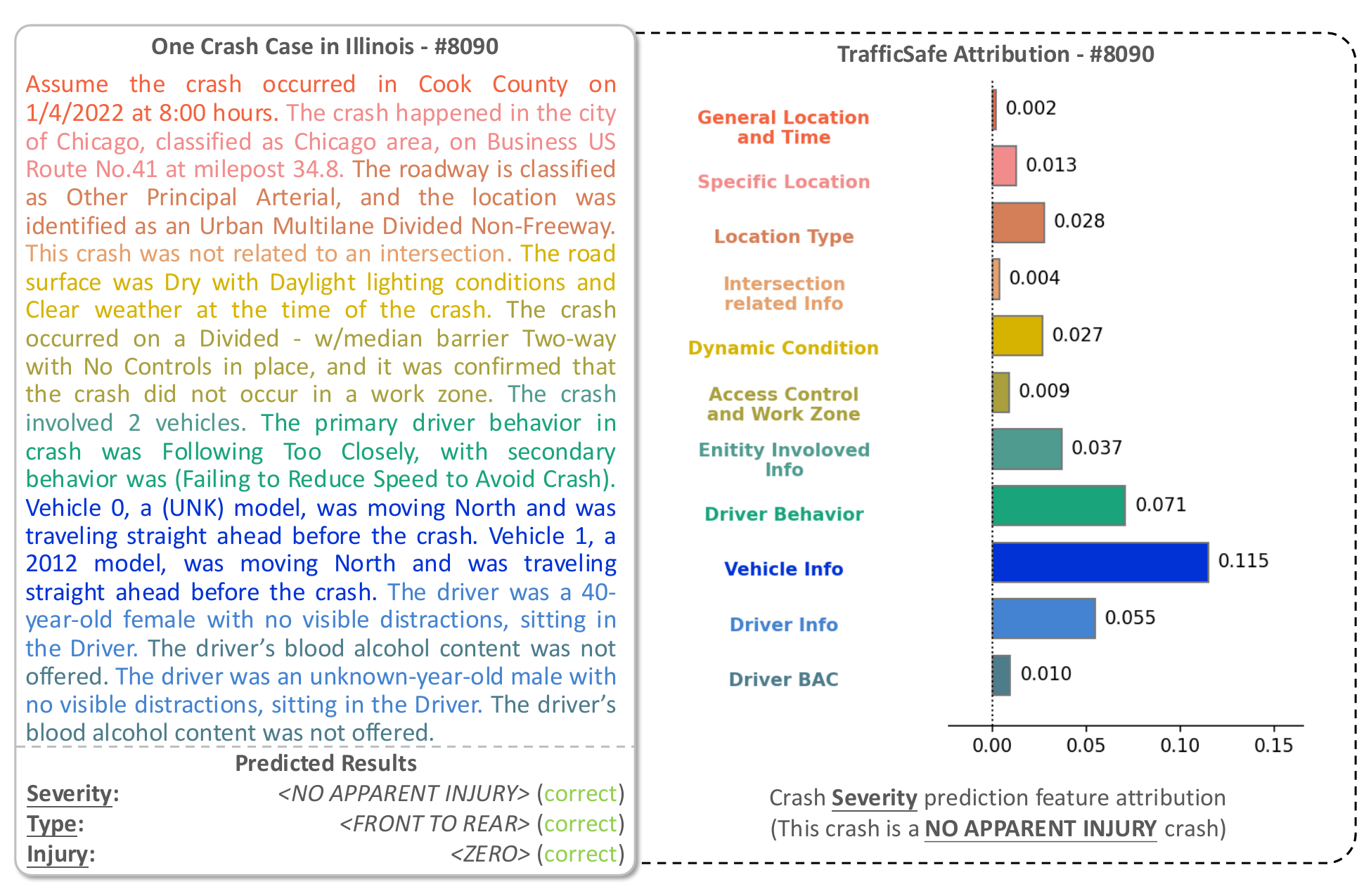}
    \vspace{-0.2cm}
    \caption{\textbf{
    One Example of Sentence-based Feature Attribution Results for A Crash Resulting in \textit{No Apparent Injury} in Illinois Dataset.}}
    \label{fig:IL_FA1}
\end{figure}

\begin{figure}[!h]
    \centering
    \stepcounter{extfig}
    \includegraphics[width=.85\linewidth]{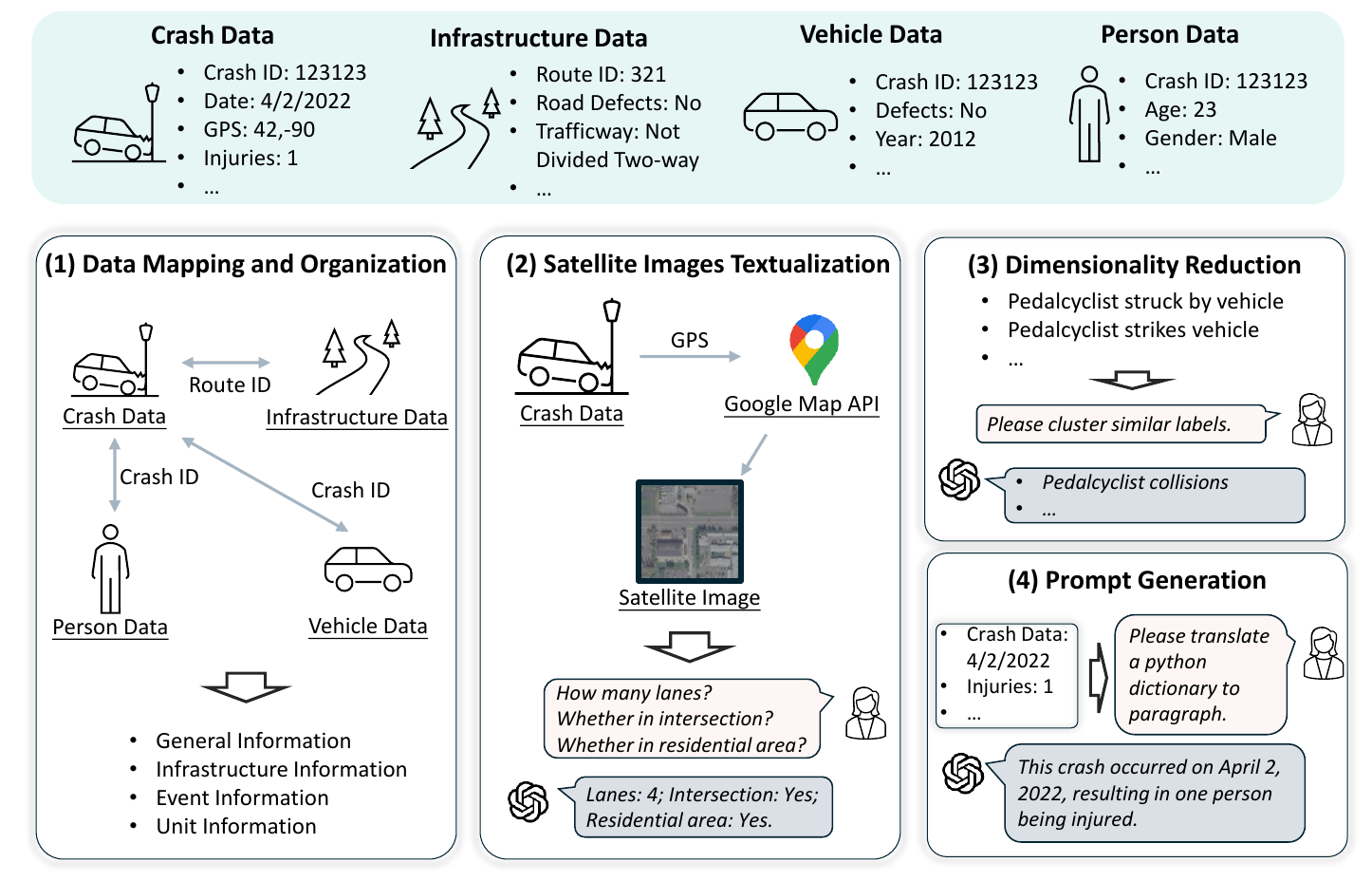}
    \caption{\textbf{Data Processing Process.} Four raw datasets from HSIS (crash, infrastructure, vehicle, and person data) are used to construct a prompt through four steps. (1) Data mapping and organization: Link the datasets and organize them into four parts: general, infrastructure, event, and unit. (2) Satellite image textualization: Retrieve satellite images via GPS coordinates using the Google Maps API, then employ GPT-4o to extract text-based information. (3) Dimensionality reduction: Combine targets with similar values using GPT-4o. (4) Prompt generation: Use the processed data from the previous steps to generate a prompt for each part.}
    \label{fig:data_processing}
\end{figure}
\newpage

\begin{figure}[!h]
    \centering
    \includegraphics[width=.85\linewidth]{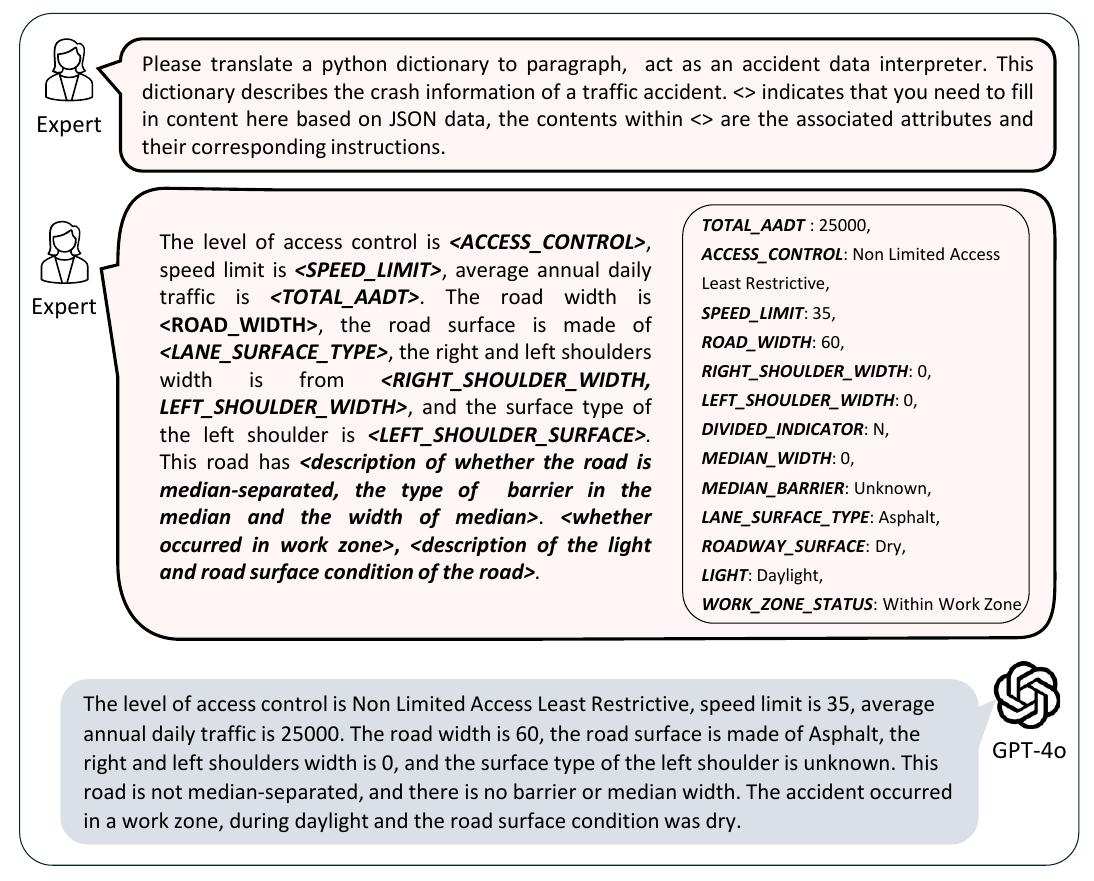}
    \caption{\textbf{AI-expert Textualization Process.} An example for the infrastructure information part of an event case in Washington dataset is shown. }
    \label{fig:ai_expert}
\end{figure}